\documentclass[journal]{IEEEtran}
\usepackage{amsmath,amsfonts}
\usepackage{algorithmic}
\usepackage{algorithm}
\usepackage{array}
\usepackage[caption=false,font=normalsize,labelfont=sf,textfont=sf]{subfig}
\usepackage{textcomp}
\usepackage{stfloats}
\usepackage{url}
\usepackage{verbatim}
\usepackage{graphicx}
\usepackage{cite}
\usepackage{makecell}
\usepackage{caption}
\usepackage{multirow}
\usepackage{xurl}
\usepackage{nccmath}

\usepackage{adjustbox}
\usepackage{amsfonts}
\usepackage{xurl}
\usepackage{relsize}
\usepackage{siunitx}
\usepackage{amsmath}
\usepackage{tablefootnote}
\usepackage{hyperref}

\usepackage{xcolor}
\definecolor{newcolor}{rgb}{.8,.349,.1}

\begin{document}

\title{Knowledge Graph Embedding with 3D Compound Geometric Transformations}


\author{Xiou Ge,
~Yun-Cheng Wang,~\IEEEmembership{Student Member,~IEEE,}
~Bin Wang,~\IEEEmembership{Member,~IEEE,}\\
~C.-C. Jay Kuo,~\IEEEmembership{Fellow,~IEEE \& ACM,}

\thanks{Xiou Ge, Yun-Cheng Wang, and C.-C. Jay Kuo are with the Ming
Hsieh Department of Electrical and Computer Engineering, University of
Southern California, Los Angeles, CA, 90089, USA. \\ Bin Wang is with National University of
Singapore, Singapore.}


}



\maketitle

\begin{abstract}

The cascade of 2D geometric transformations were exploited to model
relations between entities in a knowledge graph (KG), leading to an
effective KG embedding (KGE) model, CompoundE.  Furthermore, the
rotation in the 3D space was proposed as a new KGE model, Rotate3D, by
leveraging its non-commutative property. Inspired by CompoundE and
Rotate3D, we leverage 3D compound geometric transformations, including
translation, rotation, scaling, reflection, and shear and propose a
family of KGE models, named CompoundE3D, in this work.  CompoundE3D
allows multiple design variants to match rich underlying characteristics
of a KG. Since each variant has its own advantages on a subset of
relations, an ensemble of multiple variants can yield superior
performance. The effectiveness and flexibility of CompoundE3D are experimentally verified on four popular link prediction datasets. 

\end{abstract}


\section{Introduction}\label{3Dsec:introduction}

\IEEEPARstart{K}{nowledge} graphs (KGs) find rich applications in
knowledge management and discovery \cite{yu2017knowledge, sang2018sematyp, zeng2022toward}, recommendation systems \cite{wang2018ripplenet, zhou2020improving}, fraud
detection \cite{zhan2018loan, zhu2021intelligent}, chatbots \cite{he2017learning,ait2020kbot}, etc. KGs are directed relational graphs. They are formed by a
collection of triples in form of $(h, r, t)$, where $h$, $r$, and $t$
denote head, relation, and tail, respectively. Heads and tails are
called entities and represented by nodes while relations are links in
KGs.  KGs are often incomplete.  One critical task in knowledge graph
(KG) management is ``missing link prediction".  Knowledge graph
embedding (KGE) methods have received a lot of attention in recent years
due to their effectiveness in missing link prediction.  Many KGs such as
DBpedia \cite{auer2007dbpedia}, YAGO \cite{suchanek2007yago}, Freebase
\cite{bollacker2008freebase}, NELL \cite{carlson2010toward}, Wikidata
\cite{vrandevcic2014wikidata}, and ConceptNet \cite{speer2017conceptnet}
have been created and made publicly available for KGE model development and evaluation. 

One family of KGE models builds a high-dimensional embedding space,
where each entity is a vector. The relation is modeled by a certain
geometric manipulation such as translation and rotation. To evaluate the
likelihood of a candidate triple, the geometric manipulation associated
with the relation is applied to the head entity and then the distance
between the manipulated head and the tail is measured. The shorter the
distance, the higher likelihood of the triple.  To this end, these KGE
models are called distance-based KGEs. Examples of distance-based KGEs
include TransE \cite{bordes2013translating}, RotatE
\cite{sun2018rotate}, and PairRE \cite{chao2021pairre}. Each of them
uses a single geometric transformation to represent relations between
entities. Specifically, translation, rotation, and scaling operations
are adopted by TransE, RotatE, and PairRE, respectively. 

The above-mentioned KGE models achieve reasonably good performance in
link prediction with only a single geometric transformation. The cascade
of multiple 2D geometric transformations offers a powerful tool in image
manipulation \cite{pratt2013introduction}. This idea was exploited to
develop a new KGE model, called CompoundE, in \cite{ge2022compounde}.
TransE, RotatE and PairRE are all degenerate cases of CompoundE.  Thus,
CompoundE outperforms them in link prediction performance.  CompoundE
unifies translation, rotation, and scaling operations under one common
framework.  It has several mathematically provable properties that
facilitate the modeling of different complex relation types in KGE.  The
effectiveness of these composite operators has been successfully
demonstrated through extensive experiments and applications in
downstream tasks such as entity typing and multihop query answering in
\cite{ge2022compounde}. Furthermore, borrowed from the concept of
rotation in the 3D space, Rotate3D~\cite{gao2020rotate3d,yang2020nage,
lu2022dense} achieves more effective parameterization and endows a model
with greater modeling power than RotatE based on 2D rotation. That is,
Rotate3D can model non-commutative relations better than RotatE. 

Inspired by the success of CompoundE and Rotate3D, we wonder whether it
would be beneficial to look for compound geometric transformations in
the 3D space in the KGE model design.  Here, we extend the CompoundE
work in \cite{ge2022compounde} along three directions. First, we include
more affine operations beyond translation, rotation, and scaling such as
reflection and shear. Second, we extend these geometric transformations
from the 2D space to the 3D space and propose a family of KGE models,
CompoundE3D.  Third, CompoundE3D allows multiple design variants to
match rich underlying characteristics of a KG. Since each variant has
its own advantages on a subset of relations, an ensemble of multiple
variants can yield superior performance. The effectiveness of
CompoundE3D is experimentally verified on four popular link prediction
datasets. 

It is worthwhile to emphasize that we enhance CompoundE by addressing
two critical issues.  First, compound operations lead to numerous model
variants, and it is unclear how to determine a scoring function that
performs the best for a given dataset. Here, we propose an adapted beam search
algorithm that builds more complex scoring functions from simple but
effective ones gradually. Second, although ensemble learning is a
popular strategy, it remains under-explored when it comes to building
KGE models. In this work, we explore two ensemble strategies that
potentially boost link prediction performance and allow different
CompoundE3D variants to work together and complement each other. First,
we implement a weighted sum of different scoring functions for link
prediction.  Second, we apply unsupervised rank aggregation functions to
unify rank predictions from individual model variants. Both strategies
help boost the ranking of valid candidate entities and reduce the impact of outliers. 

The major contributions of this work are summarized below.
\begin{itemize}
\item We examine affine operations in the 3D space, instead of the 2D
space, to allow more versatile relation representations. Besides
translation, rotation, and scaling used in CompoundE, we include reflection
and shear transformations which allow an even larger design space. 
\item We propose an adapted beam search algorithm to discover better model
variants. Such a procedure avoids unnecessary exploration of poor
variants but zooms into more effective ones to strike a good balance
between model complexity and prediction performance. 
\item We analyze the properties of each operation and its advantage in
modeling different relations. Our analysis is backed by empirical
results on four datasets. 
\item To reduce errors of an individual model variant and boost the
overall link prediction performance, we aggregate decisions from
different variants with two approaches; namely, the sum of weighted
distances and rank fusion. 
\end{itemize}

The rest of this paper is organized as follows.  A brief review of
related work is provided in Sec. \ref{3Dsec:related_work}.  The design
methodology of CompoundE3D is explained and the decision ensemble of
multiple model variants is elaborated in Sec. \ref{3Dsec:method}.
Experimental results and performance benchmarking with previous work are
presented in Sec.  \ref{3Dsec:experiments}.  Finally, concluding remarks
and future research directions are given in Sec. \ref{3Dsec:conclusion}. 

\section{Related Work}\label{3Dsec:related_work}

\subsection{Knowledge Graph Embedding (KGE) Models}

A large number of distance-based KGE models are derived by treating
relations as certain transformations.  They are briefly reviewed below. 

\subsubsection{2D Geometric Transformations}

Quite a few KGE models are inspired by 2D geometric transformations such
as translation, rotation, and scaling in the 2D plane.  TransE
\cite{bordes2013translating} models the relation as a translation
between head and tail entities. This simple model is not able to model
symmetric relations effectively. RotatE \cite{sun2018rotate} treats
relations as certain rotations in the complex space, which works well
for symmetric relations. Furthermore, RotatE introduces a
self-adversarial negative sampling loss that improves distance-based KGE
model performance significantly. PairRE \cite{chao2021pairre} models
relations with the scaling operation to allow variable margins. This is
helpful in encoding complex relations. The unitary constraint on entity
embedding in PairRE is also effective in practice. CompoundE
\cite{ge2022compounde} adopts compound geometric transformations,
including translation, rotation, and scaling, to model different
relations.  It offers a superior KGE model without increasing the
overall complexity much. 

\subsubsection{Advanced Transformations}

NagE \cite{yang2020nage} introduces generic group theory to the design
of KGE models and gives a generic recipe for their construction. QuatE
\cite{zhang2019quaternion} extends the KGE design to the Quaternion
space which enables more compact interactions between entities and
relations while introducing more degree of freedom.  To model
non-commutativeness in relation composition more effectively, both
RotatE3D \cite{gao2020rotate3d} and DensE \cite{lu2022dense} leverage
quaternion rotations but in different forms. ROTH \cite{chami2020low}
adopts the hyperbolic curvature to capture the hierarchical structure in
KGs. On the other hand, it is questioned in \cite{wang2021hyperbolic}
whether the introduction of hyperbolic geometry in KGE is necessary. 

\subsection{Classification-based Models}

Another family of models is built by classifying an unseen triple into
``valid" (or positive) and ``invalid" (or negative) two classes and then
using the soft decision to measure the likelihood of the triple. 

\subsubsection{Simple Neural Networks}

A multilayer perceptron (MLP) network \cite{dong2014knowledge} is used
to measure the likelihood of unseen triples for link prediction.  The
neural tensor network (NTN) \cite{socher2013reasoning} adopts a bilinear
tensor neural layer to model interactions between entities and relations
of triples. ConvE \cite{dettmers2018convolutional} stacks head entities
and relations, reshapes them to 2D arrays, and uses the convolutional
neural network (CNN) to extract the information from them. The resulting
feature map interacts with tail entities through dot products.  R-GCN
\cite{schlichtkrull2018modeling} uses the graph convolutional network
(GCN) with relation-specific weights to obtain entity representations,
which are subsequently fed to DistMult \cite{yang2014embedding} for link
prediction. Despite its potential of handling the inductive setting, its
performance is not on par with the embedding based approach. 

\subsubsection{Advanced Neural Networks}

KG-BERT \cite{yao2019kg} uses the pretrained language model, BERT
\cite{devlin2018bert}, to obtain the entity representation from textual
descriptions (rather than from KG links).  However, its inference time
is much longer compared to embedding-based models.  SimKGC
\cite{wang2022simkgc} improves transformer-based classification methods
by constructing contrastive pairs. It uses BERT to estimate the semantic
similarity and treats triples of higher similarity score as positive
sample pairs, and vice versa. However, its performance is sensitive to
the language model quality, and its required computational resource is
high. 

\subsubsection{Lightweight Classification Model}

KGBoost \cite{wang2022kgboost} proposes a novel negative sampling
scheme, and uses the XGBoost \cite{chen2016xgboost} classifier for link
prediction.  Inspired by the Discriminant Feature Learning (DFT)
\cite{yang2022supervised, kuo2022green} that extracts most
discriminative features from trained embeddings, GreenKGC \cite{wang2022greenkgc} is a
lightweight and modularized classification method that trains a binary
classifier to classify unseen triples. 

\subsection{Advanced Relation Modeling}

Special techniques have been developed to model complex relations.  For
example, to model relations such as 1-to-N, N-to-1, and N-to-N
effectively, TransH \cite{wang2014knowledge} projects the embedded
entity space into relation-specific hyper-planes. TransR
\cite{lin2015learning} learns a relation-specific projection that maps
entity vectors to a certain relation space. TransD
\cite{ji2015knowledge} derives dynamic mapping based on relation and
entity projection vectors.  TranSparse \cite{ji2016knowledge} enforces
the relation projection matrix to be sparse. Recently, many KGE models
including X+AT \cite{yang2021improving}, SFBR \cite{liang2021semantic},
and STaR \cite{li2022star} apply translation and scaling operations to both
distance-based and semantic-matching-based \cite{wang2017knowledge} models to improve the
performance gain. The inclusion of translation is proven to be effective
in improving KGEs in the Quaternion space such as DualE
\cite{cao2021dual}, BiQUE \cite{guo2021bique}. ReflectE
\cite{zhang2022knowledge} models each relation as a normal vector of a
hyper-plane that reflects entity vectors. It can be used to model
symmetric and inverse relations well. So far, the cascade of various
affine operations is a natural yet unexplored idea to pursue. 

\subsection{Model Ensembles}

Although ensemble learning is a prevailing strategy in machine learning,
it remains under-explored for knowledge graph completion. Link
prediction evaluation is essentially a ranking problem. It is desired to
optimize an ensemble decision so that valid triples get ranked higher than invalid ones among all candidates. Rank aggregation is a classical problem in
information retrieval. Both supervised methods \cite{cao2007learning,
chen2016xgboost} and unsupervised methods
\cite{klementiev2008unsupervised, cormack2009reciprocal} have been
studied.  Since the ground truth ranking in KG's link prediction is not
available (except the top-1 triple), the unsupervised setting is more
relevant. Yet, the use of rank aggregation to boost link prediction
performance has received limited attention. Several examples are given
below.  KEnS \cite{chen2020multilingual} performs ensemble inference to
combine predictions from multiple language-based KGEs for multilingual
knowledge graph completion.  AutoSF \cite{zhang2020autosf} develops an
algorithm to search for the best scoring functions from multiple
semantic matching models.  The ensemble of multiple identical
low-dimensional KGE models is adopted in \cite{xu2021multiple} to boost
the link prediction performance. Recently, DuEL \cite{joshi2022ensemble}
treats link prediction as a classification problem and aggregates binary
decisions from several different classifiers using unsupervised
techniques. 

\begin{figure*}[ht!]
\centering
\subfloat[Translation\label{fig:translate}]
{\includegraphics[width=0.33\textwidth]{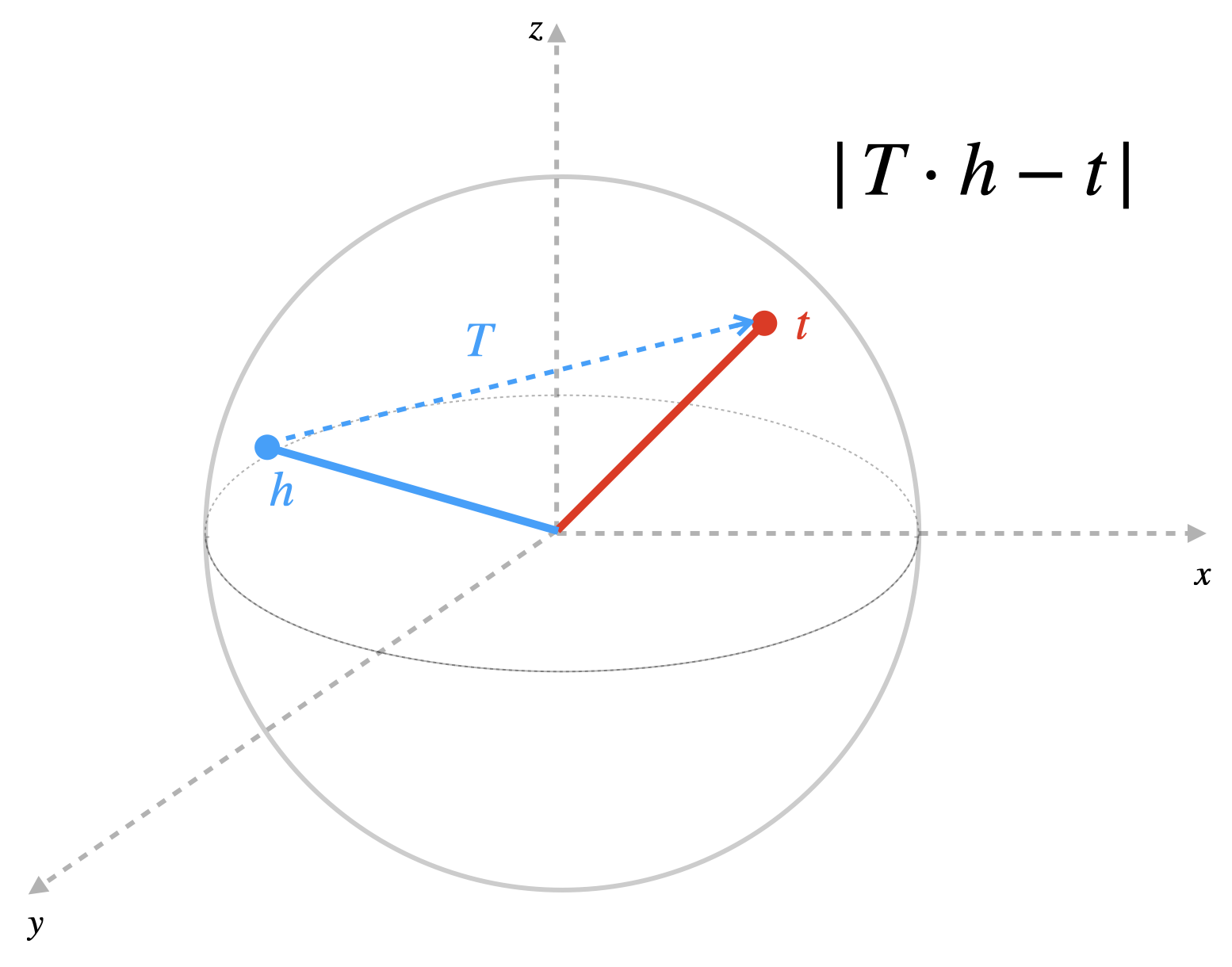}} 
\subfloat[Scaling\label{fig:scaling}]
{\includegraphics[width=0.33\textwidth]{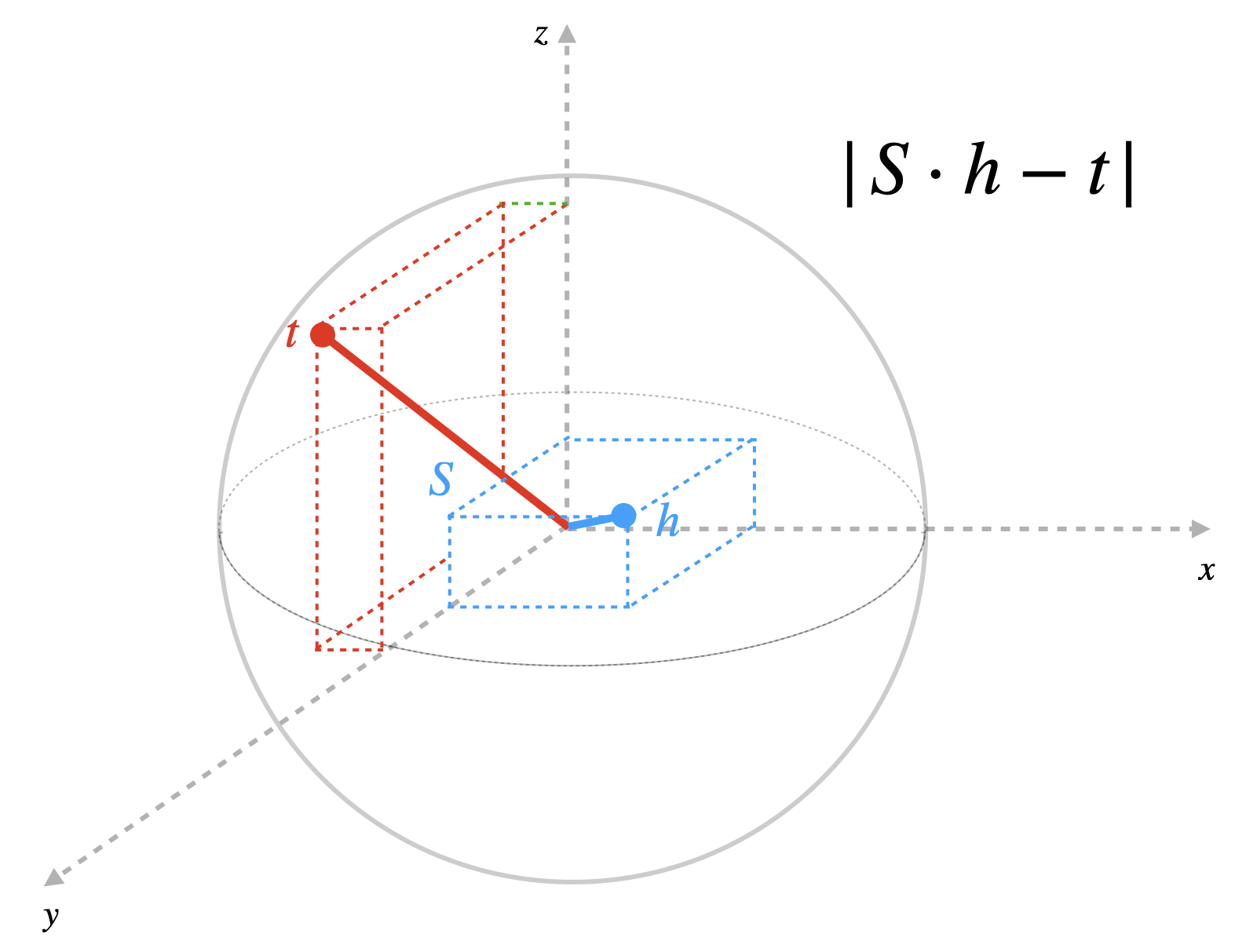}}
\subfloat[Rotation\label{fig:rotation}]
{\includegraphics[width=0.33\textwidth]{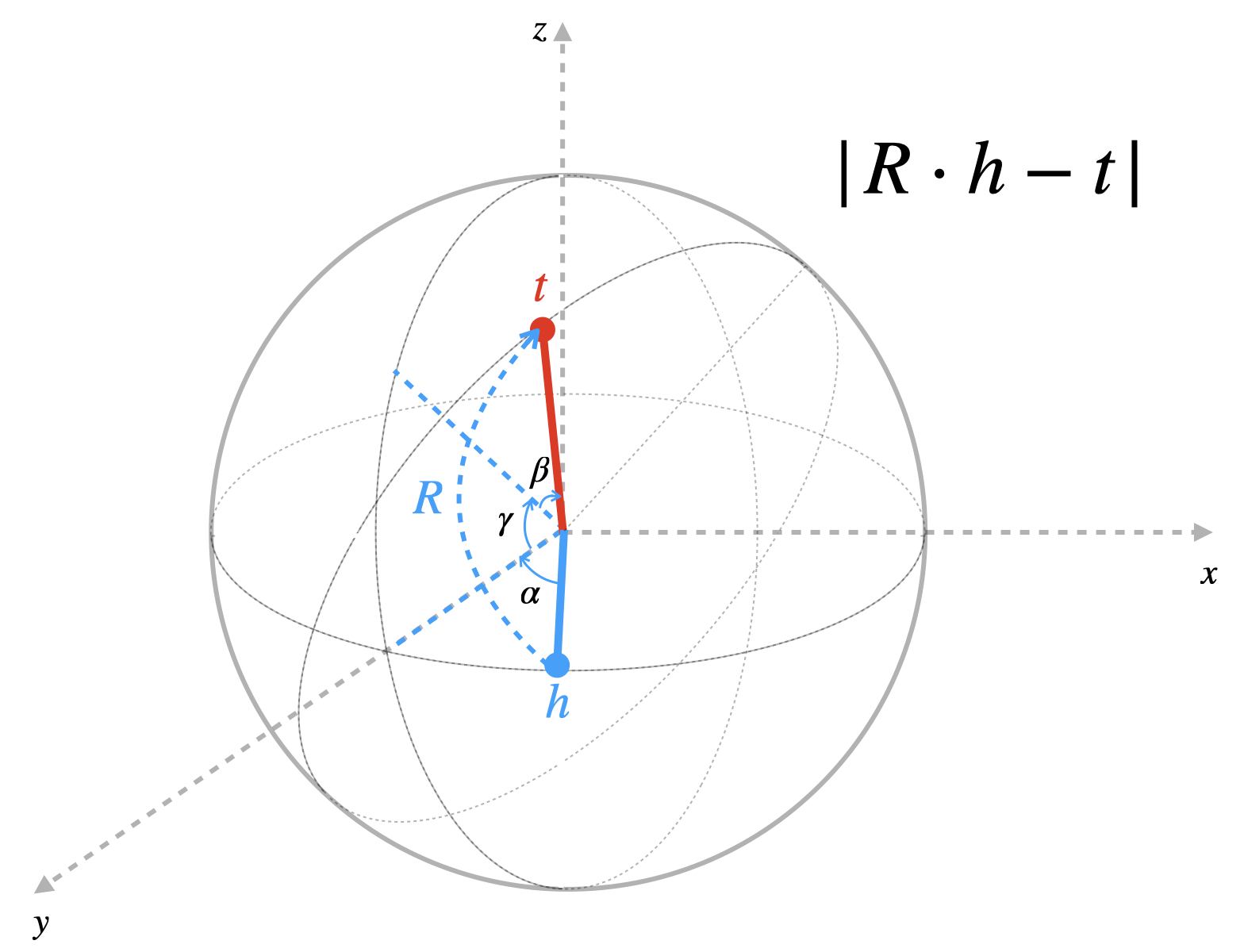}} \\
\subfloat[Reflection\label{fig:reflection}]
{\includegraphics[width=0.33\textwidth]{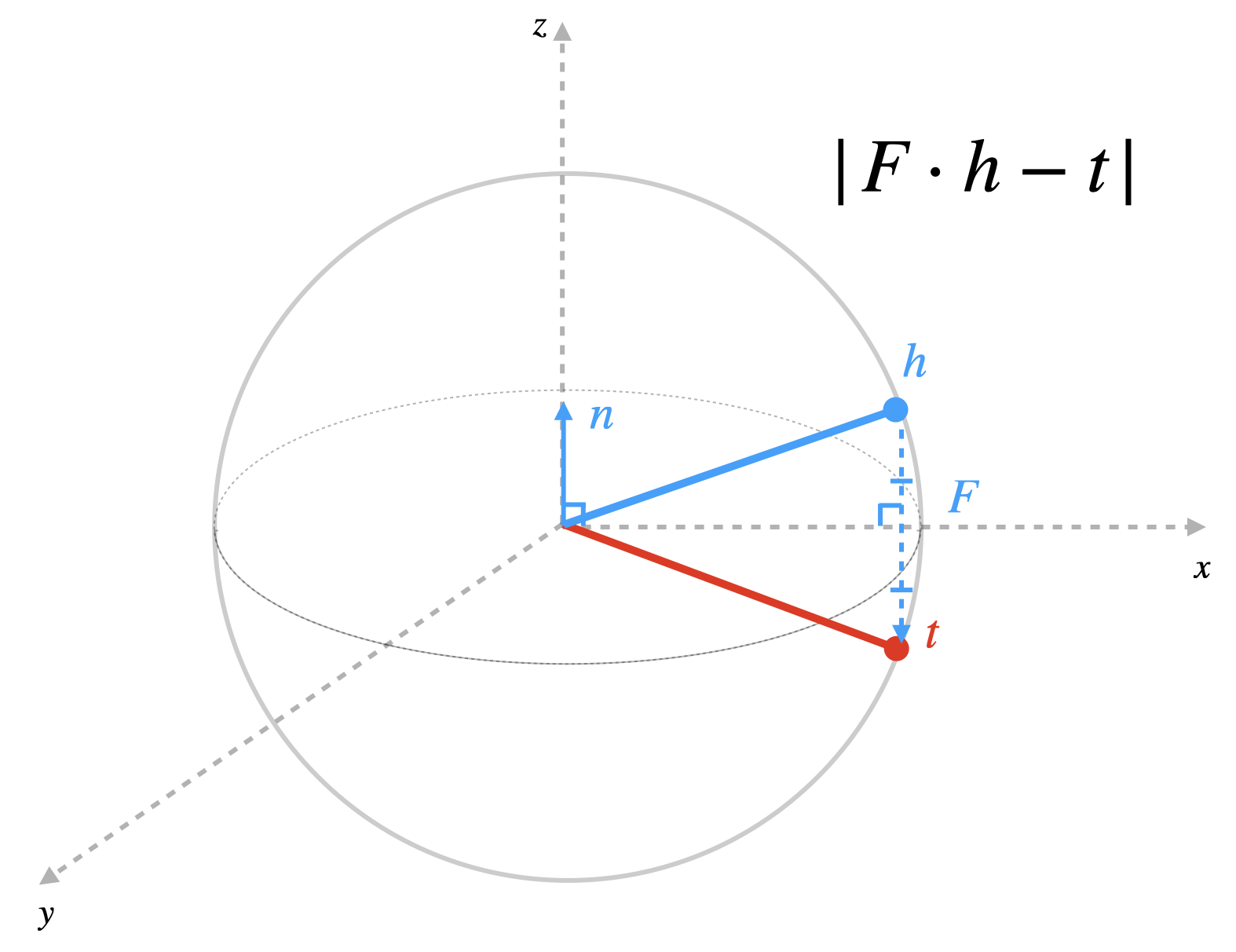}} 
\subfloat[Shear\label{fig:shear}]
{\includegraphics[width=0.33\textwidth]{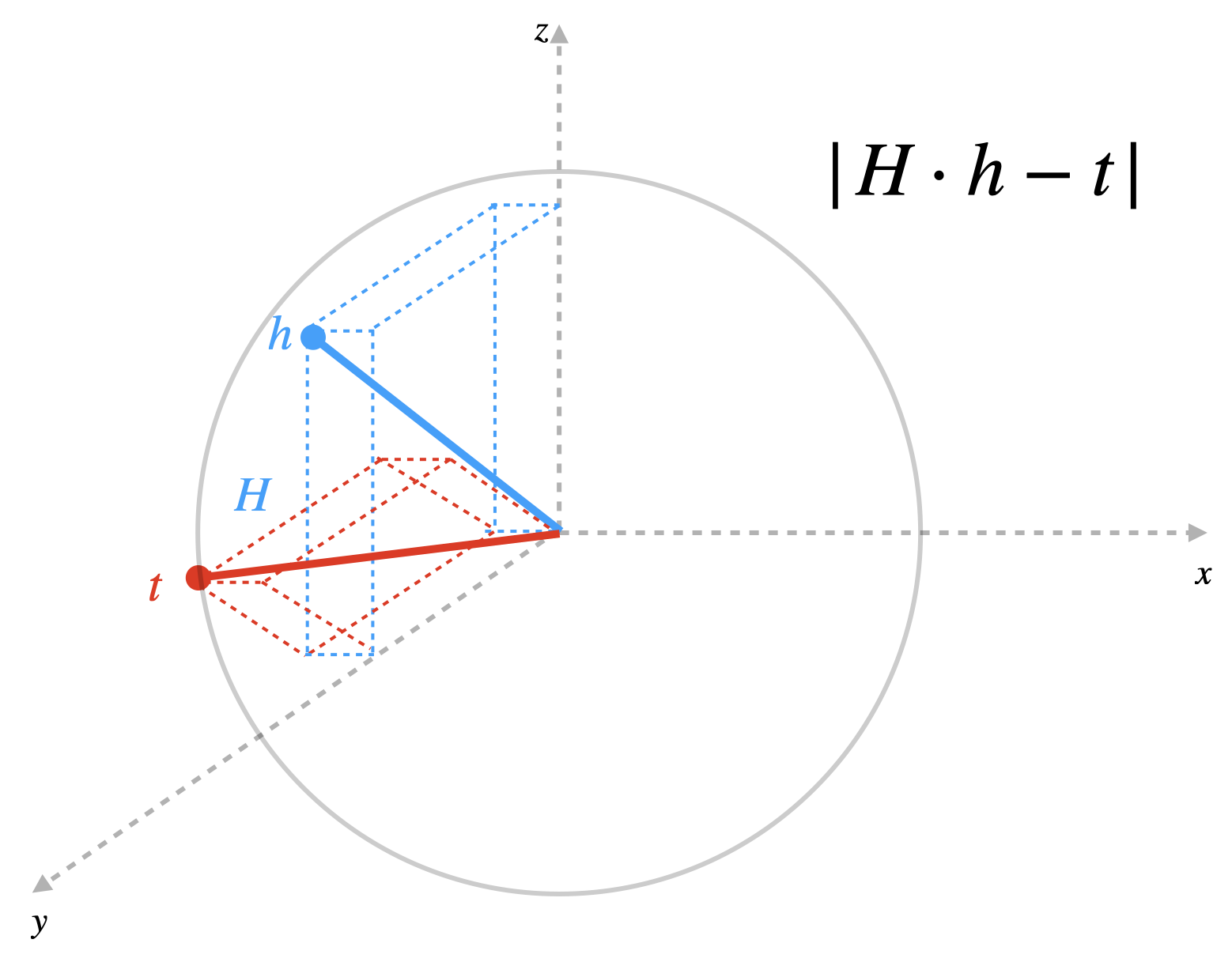}}
\subfloat[Compound\label{fig:compound}]
{\includegraphics[width=0.33\textwidth]{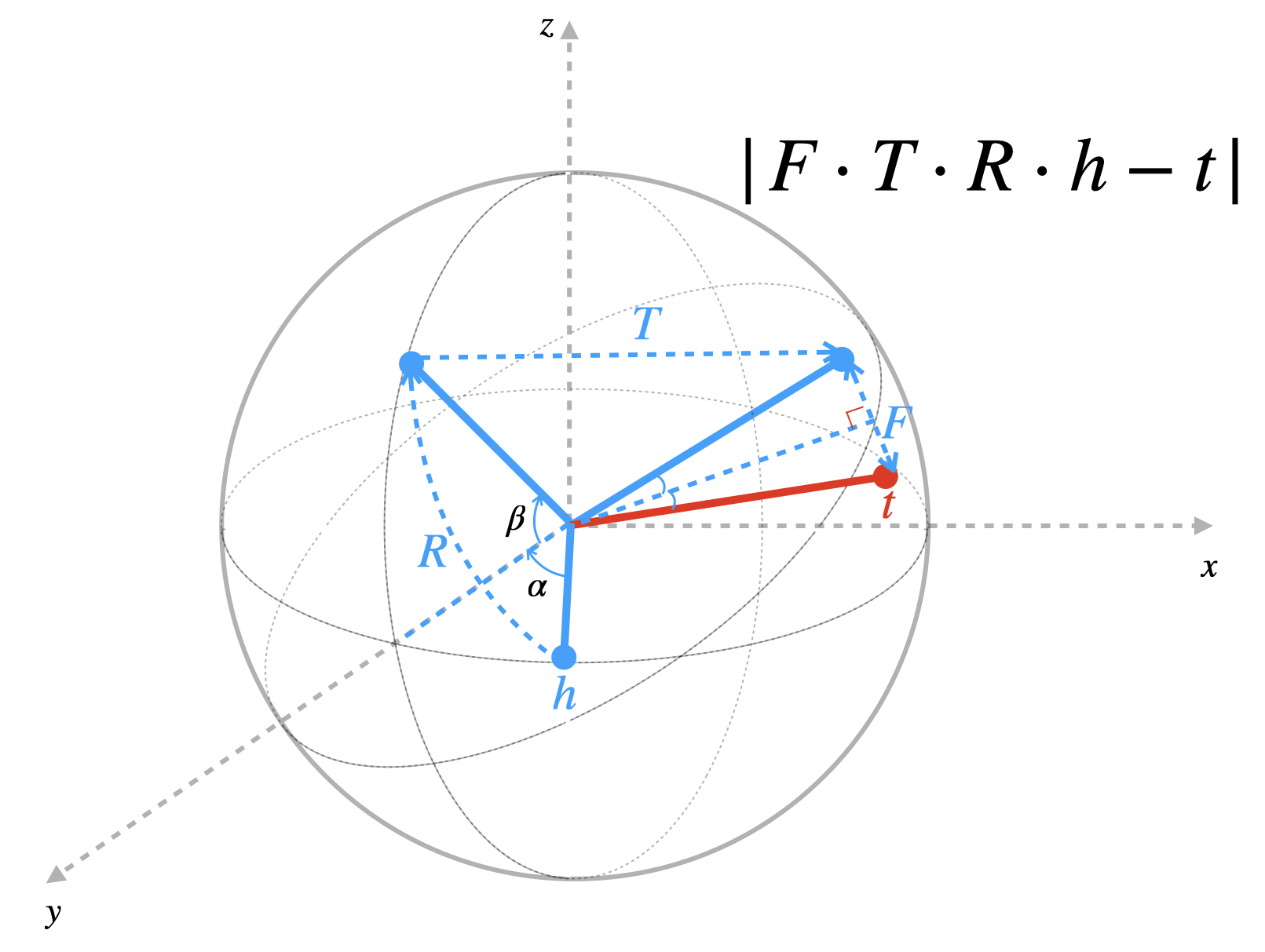}} 
\caption{Composing different geometric operations in the 3D
subspace.}\label{fig:diff_operation}
\end{figure*}

\section{Proposed Method}\label{3Dsec:method}

\subsection{CompoundE3D}

In this work, we use 3D affine transformations, including Translation,
Scaling, Rotation, Reflection, and Shear as illustrated in
\ref{fig:diff_operation}, to model different relations in KGs. This
large set of transformation operators offer immense flexibility in the
KGE design against different characteristics of KG datasets.  Below, we
formally define each of the 3D affine operators in homogeneous
coordinates. 

\subsubsection{Translation} Component $\mathbf{T}\in\mathbf{SE}(3)$,
illustrated by Fig. \ref{fig:translate}, is defined as
\begin{equation}
    \mathbf{T} = \begin{bmatrix}
    1 & 0 & 0 & v_x \\
    0 & 1 & 0 & v_y \\
    0 & 0 & 1 & v_z \\
    0 & 0 & 0 & 1
  \end{bmatrix},
\end{equation}

\subsubsection{Scaling} Component $\mathbf{S}\in\mathbf{Aff}(3)$,
illustrated by Fig. \ref{fig:scaling}, is defined as
\begin{equation}
    \mathbf{S} = \begin{bmatrix}
    s_x & 0 & 0 & 0 \\
    0 & s_y & 0 & 0 \\
    0 & 0 & s_z & 0 \\
    0 & 0 & 0 & 1
  \end{bmatrix},
\end{equation}

\subsubsection{Rotation} Component $\mathbf{R}\in\mathbf{SO}(3)$,
illustrated by Fig. \ref{fig:rotation}, is defined as
\begin{equation}
    \mathbf{R} = \mathbf{R}_z(\alpha)\mathbf{R}_y(\beta)\mathbf{R}_x(\gamma) = 
    \begin{bmatrix}
    a & b & c & 0 \\
    d & e & f & 0 \\
    g & h & i & 0 \\
    0 & 0 & 0 & 1
  \end{bmatrix},
\end{equation}
where
\begin{equation}
    \begin{aligned}
        a &= \cos(\alpha)\cos(\beta), \\
        b &= \cos(\alpha)\sin(\beta)\sin(\gamma)-\sin(\alpha)\cos(\gamma), \\
        c &= \cos(\alpha)\sin(\beta)\cos(\gamma)+\sin(\alpha)\sin(\gamma), \\
        d &= \sin(\alpha)\cos(\beta), \\
        e &= \sin(\alpha)\sin(\beta)\sin(\gamma)+\cos(\alpha)\cos(\gamma), \\
        f &= \sin(\alpha)\sin(\beta)\cos(\gamma)-\cos(\alpha)\sin(\gamma), \\
        g &= -\sin(\beta), \\
        h &= \cos(\beta)\sin(\gamma), \\
        i &= \cos(\beta)\cos(\gamma). \\
    \end{aligned}
\end{equation}
This general 3D rotation operator is the result of compounding yaw,
pitch, and roll rotations. They are, respectively, defined as
\begin{itemize}
\item Yaw rotation component:
\begin{equation}
    \mathbf{R}_z(\alpha) = \begin{bmatrix}
    \cos(\alpha) & -\sin(\alpha) & 0 & 0 \\
    \sin(\alpha) & \cos(\alpha) & 0 & 0 \\
    0 & 0 & 1 & 0 \\
    0 & 0 & 0 & 1
  \end{bmatrix},
\end{equation}
\item Pitch rotation component:
\begin{equation}
    \mathbf{R}_y(\beta) = \begin{bmatrix}
    \cos(\beta) & 0 & -\sin(\beta) & 0 \\
    0 & 1 & 0 & 0 \\
    \sin(\beta) & 0 & \cos(\beta) & 0 \\
    0 & 0 & 0 & 1
  \end{bmatrix},
\end{equation}
\item Roll rotation component:
\begin{equation}
    \mathbf{R}_x(\gamma) = \begin{bmatrix}
    1 & 0 & 0 & 0 \\
    0 & \cos(\gamma) & -\sin(\gamma) & 0 \\
    0 & \sin(\gamma) & \cos(\gamma) & 0 \\
    0 & 0 & 0 & 1
  \end{bmatrix}.
\end{equation}
\end{itemize}

\subsubsection{Reflection} Component $\mathbf{F}\in\mathbf{SO}(3)$,
illustrated by Fig. \ref{fig:reflection}, is defined as
\begin{equation}
    \mathbf{F} = \begin{bmatrix} 
    1 - 2n_x^2 & -2n_x n_y & -2n_x n_z & 0 \\
    -2n_x n_y & 1 - 2 n_y^2 & -2n_y n_z & 0 \\
    -2n_x n_z & -2n_y n_z & 1 - 2n_z^2 & 0 \\
    0 & 0 & 0 & 1 
    \end{bmatrix}.
\end{equation}
The above expression is derive from the Householder reflection, $\mathbf{F =
I - 2nn^T}$. In the 3D space, $\mathbf{n}$ is a 3-D unit vector that is
perpendicular to the reflecting hyper-plane, $\mathbf{n} = [n_x, n_y,
n_z]$. 

\subsubsection{Shear} Component $\mathbf{H}\in\mathbf{Aff}(3)$,
illustrated by Fig. \ref{fig:shear}, is defined as
\begin{equation}
    \mathbf{H} = \mathbf{H}_{yz} \mathbf{H}_{xz} \mathbf{H}_{xy} = \begin{bmatrix}
    1 & \text{Sh}^y_x & \text{Sh}^z_x & 0 \\
    \text{Sh}^x_y & 1 & \text{Sh}^z_y & 0 \\
    \text{Sh}^x_z & \text{Sh}^y_z & 1 & 0 \\
    0 & 0 & 0 & 1
  \end{bmatrix}.
\end{equation}
The shear operator is the result of compounding 3 operators:
$\mathbf{H}_{yz}$, $\mathbf{H}_{xz}$, and $\mathbf{H}_{xy}$ 
They are mathematically defined as
\begin{eqnarray}
\mathbf{H}_{yz} & = & \begin{bmatrix}
    1 & 0 & 0 & 0 \\
    \text{Sh}^x_y & 1 & 0 & 0 \\
    \text{Sh}^x_z & 0 & 1 & 0 \\
    0 & 0 & 0 & 1 \\
\end{bmatrix}, \\
\mathbf{H}_{xz} & = & \begin{bmatrix}
    1 & \text{Sh}^y_x & 0 & 0 \\
    0 & 1 & 0 & 0 \\
    0 & \text{Sh}^y_z & 1 & 0 \\
    0 & 0 & 0 & 1 \\
\end{bmatrix}, \\
\mathbf{H}_{xy} & = & \begin{bmatrix}
    1 & 0 & \text{Sh}^z_x & 0 \\
    0 & 1 & \text{Sh}^z_y & 0 \\
    0 & 0 & 1 & 0 \\
    0 & 0 & 0 & 1 \\
\end{bmatrix}.
\end{eqnarray}
Matrix $\mathbf{H}_{yz}$ has a physical meaning - the shear
transformation that shifts the $y$- and $z$- components by a factor of
the $x$ component.  Similar physical interpretations are applied to
$\mathbf{H}_{xz}$ and $\mathbf{H}_{xy}$. 

The above transformations can be cascaded to yield a compound operator; e.g.,
\begin{equation}
\mathbf{O = T\cdot S\cdot R\cdot F\cdot H},
\end{equation}
In the actual implementation, we use the operator's representation in regular Cartesian coordinate instead of the homogeneous coordinate. Furthermore, a high-dimensional relation operator can
be represented as a block diagonal matrix in the form of
\begin{equation}\label{eq:relation_operator}
    \mathbf{M_r = \textbf{diag}(O_{r,1}, O_{r,2}, \dots, O_{r,n})},
\end{equation}
where $\mathbf{O_{r,i}}$ is the compound operator at the $i$-th stage. 

We can define the following three scoring functions
for CompoundE3D:
\begin{itemize}
\item CompoundE3D-Head
\begin{equation}
f_r^{(h)} (h,t) = \|\mathbf{M_r\cdot h - t }\|,
\end{equation}
\item CompoundE3D-Tail
\begin{equation}
f_r^{(t)}(h,t) = \|\mathbf{h - \hat{M}_r\cdot t }\|,
\end{equation}
\item CompoundE3D-Complete
\begin{equation}
f_r^{(h,t)}(h,t) = \|\mathbf{M_r\cdot h - \hat{M}_r\cdot t }\|,
\end{equation}
\end{itemize}
where $\mathbf{h}$ and $\mathbf{t}$ denote head and tail entity
embeddings, and $\mathbf{M_r}$ and $\mathbf{\hat{M}_r}$ denote the
relation-specific operators that operate on head and tail entities,
respectively. 

Generally speaking, we have five different affine operations available
to use, i.e.  translation, scaling, rotation, reflection, and shear.
Each operator can be applied to 1) head entity, 2) tail entity, or 3)
both head and tail.  Hence, we have in total 15 different ways of
applying operators at each stage. All these possible choices are called
CompoundE3D variants. For a given KG dataset, there is a huge search
space in finding the optimal CompoundE3D variant. It is essential to
develop a simple yet effective mechanism to find a variant that gives
the best performance under a certain complexity constraint.

\subsection{Beam Search for Best CompoundE3D Variant}\label{subsec:beam_search}

In this subsection, we present a beam search algorithm to find the
optimal CompoundE3D variant.  For the $i$-th stage, the set of all
operator pairs that can be applied at a certain step is
\begin{equation}
\begin{aligned}
    \mathbf{P} \in\{ & \mathbf{ (T, I), (S, I), (R, I), (F, I), (H, I),} \\ 
    & \mathbf{(I, \hat{T}), (I, \hat{S}), (I, \hat{R}), (I, \hat{F}), (I, \hat{H}),} \\ 
    &\mathbf{(T, \hat{T}), (S, \hat{S}), (R, \hat{R}), (F, \hat{F}), (H, \hat{H})}\},
\end{aligned}
\end{equation}
where $\mathbf{I}$ is the identity operator.  First, we apply all operator pairs
in $\mathbf{P}$ and calculate scoring functions for all intermediate
variants. Each variant is optimized with $l$ iterations using the
training set and its performance is evaluated on the validation dataset.
Then, we choose the top-$k$ best-performing variants as starting points
for further exploration in the next step. The same process is repeated
until one of terminating conditions is triggered. Afterward, we proceed
to the $(i+1)$-th stage. The whole search is completed after the final stage
is reached. The total number of stages is a user selected hyper-parameter.

The beam search process in building more complex KGE models from simpler
ones is described in Algorithm \ref{alg:beam_search}. Additional comments 
are given below.
\begin{itemize}
\item We initialize the algorithm by setting up a loop to iterate over
the set, $\mathbf{P}$, of all possible operator combinations to train
and evaluate them and find the top-$k$ variants as starting points. 
\item In the next loop, we have two stopping criteria to terminate the
beam search: 1) $\#$ operators $> \lambda$, meaning that we stop the
search when the number of operators exceeds the upper bound $\lambda$; 2)
$\frac{\Delta \text{MRR}}{\Delta \text{Param}}<\gamma$, meaning that the
ratio of increase in MRR versus the increase in free parameters fall
below the threshold $\gamma$, and it is no longer worthwhile to increase
the model complexity for the marginal gain in model performance. 
\item $\mathbf{P} \times \mathbf{W}$ denotes the Cartesian product
between the operator pairs set $\mathbf{P}$ and top-$k$ variants set
$\mathbf{W}$ from the last step while $\dot{f}^{i-1}_r(h, t) \lhd
(\mathbf{M}^i, \mathbf{\hat{M}}^i)$ denotes applying the operator pair
$(\mathbf{M}^i, \mathbf{\hat{M}}^i)$ to previous optimal scoring
function $\dot{f}^{i-1}_r(h, t)$. 
\item For example, if $\dot{f}^{i-1}_r(h, t) = \|\mathbf{R\cdot h- t}\|$ and
$(\mathbf{M}^i, \mathbf{\hat{M}}^i) = (\mathbf{S}, \mathbf{\hat{S}})$,
then $\Tilde{f}_r^i(h, t) = \|\mathbf{S\cdot R\cdot h- \hat{S}\cdot t}\|$. 
\item After the loop terminates due to any terminating condition is
triggered, we select the top-1 performing variant from the explored
variants set, $\mathbf{W}$, as the best choice. 
\end{itemize}

\begin{algorithm}
\caption{Beam Search for Best CompoundE3D Variant}\label{alg:beam_search}
\begin{algorithmic}
\STATE \textbf{initialize} $i \gets 1, \mathbf{U} \gets \{\}$
\FOR{$(\mathbf{M}^i, \mathbf{\hat{M}}^i)\in \mathbf{P}$}
    \STATE $\Tilde{f}_r^i(h, t) \gets \|\mathbf{M}^i
      \cdot\mathbf{h}-\mathbf{\hat{M}}^i \cdot\mathbf{t}\|$;
    \STATE train $\Tilde{f}_r(h, t)$ for $l$ iterations;
    \STATE MRR $\gets$ evaluate $\Tilde{f}_r^i(h, t)$ with valid set; 
    \STATE $\mathbf{U}$.insert($\{\text{MRR}, \Tilde{f}_r^i(h, t)\}$);
\ENDFOR
\STATE $\mathbf{W} \gets$ top-$k$ variants from $\mathbf{U}$
\STATE $i \gets i+1$
\STATE $\Delta \text{MRR} \gets \gamma$, $\Delta \text{Param} \gets 1$
\WHILE{$\#$ operators $< \lambda$ \textbf{and} $\max \frac{\Delta 
\text{MRR}}{\Delta \text{Param}}\geq\gamma$}
    \STATE \textbf{initialize} $\mathbf{V}\gets \{\}$
    \FOR{\{$(\mathbf{M}^i, \mathbf{\hat{M}}^i), \dot{f}^{i-1}_r(h, t)\}\in 
        \mathbf{P} \times \mathbf{W}$}
        \STATE $\Tilde{f}_r^i(h, t) \gets \dot{f}^{i-1}_r(h, t) \lhd 
        (\mathbf{M}^i, \mathbf{\hat{M}}^i)$;
        \STATE train $\Tilde{f}_r^i(h, t)$ for $l$ iterations;
        \STATE evaluate $\Tilde{f}_r^i(h, t)$ with valid set;
        \STATE $\Delta$MRR $\gets$ $\Tilde{f}_r^i(h, t)$ MRR$-\dot{f}_r^{i-1}(h, t)$ MRR;
        \STATE $\Delta$Param $\gets$ $\Tilde{f}_r^i(h, t)$ Param$-\dot{f}_r^{i-1}(h, t)$ Param;
        \STATE $\mathbf{V}$.insert(MRR, $\Delta$MRR, $\Delta$Param, $\Tilde{f}_r^i(h, t)$);
    \ENDFOR
    \STATE $\mathbf{W} \gets$ top-$k$ variants from $\mathbf{V}$;
\ENDWHILE
\STATE ${f}^{*}_r(h, t) \gets$ best variant from $\mathbf{W}$;
\end{algorithmic}
\end{algorithm}

\begin{figure}[ht!]
\centering
\includegraphics[width=\columnwidth]{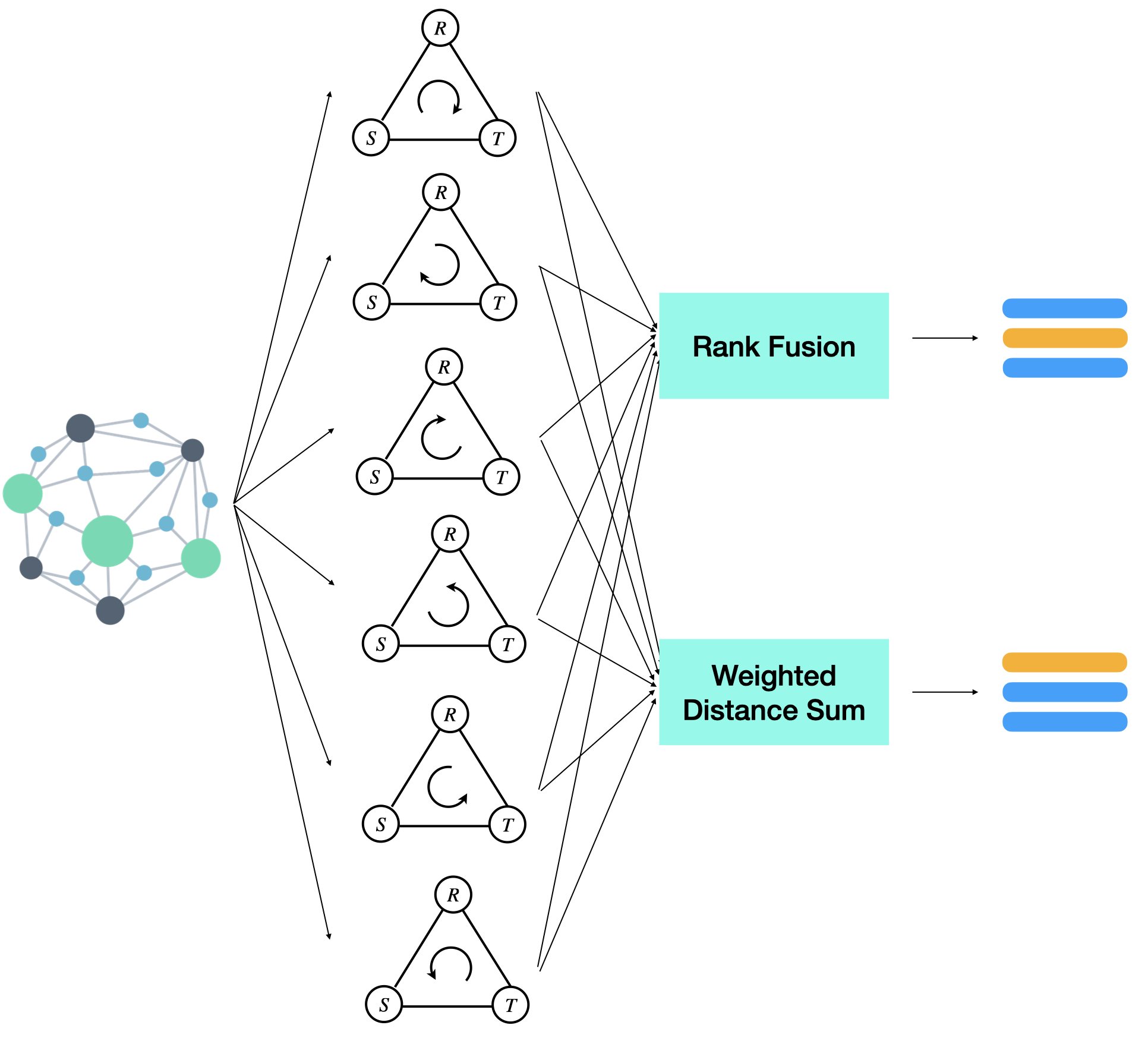}
\caption{The ensemble of multiple CompoundE3D variants.}\label{ensemble}
\end{figure}

\subsection{Model Ensembles}\label{subsec:fusion}

\subsubsection{Weighted-Distances-Sum (WDS) Strategy}

We choose the top-$k$ performing CompoundE3D variants and conduct a
weighted average of their predicted scores. The following three weighting
schemes are considered. 
\begin{itemize}
\item {\bf Uniform Weights.} This scheme takes an equal weight
of selected $k$ variants as
\begin{equation}
\hat{f}_r(h,t) = \frac{1}{k}\sum_{i=1}^k f_r^i(h,t),
\end{equation}
where $f_r^i(h,t)$ is the scoring function for the $i$-th variant.
\item {\bf Geometric Weights.} This scheme sorts the variants based on
their MRR performance on the validation dataset in a descending order 
and assign weight $\lambda^k$, $0 < \lambda < 1$, to the $k$-th variant.
That is, 
\begin{equation}
\hat{f}_r(h,t) = \frac{1}{\sum_{i=1}^k \lambda^k} \sum_{i=1}^k \lambda^k 
f_r^i(h,t).
\end{equation}
Since $\lambda^k > \lambda^{k+1}$, we assign a higher weight to a better
performer in computing the aggregated distance. 
\item {\bf Learnable Weights.} This scheme trains a set of learnable
weights, $w_i > 0$, based on the training dataset to minimize the
following weighted score:
\begin{equation}
\hat{f}_r(h,t) = \frac{1}{\sum_{i=1}^k w_i} \sum_{i=1}^k w_i f_r^i(h,t).
\end{equation}
The learnable weights are implemented as parameters in the optimization
process under the same learning rate and the optimizer in finding the
best variants. 
\end{itemize}
For each relation, we compare the three weight schemes and choose the
one that offers the best performance. 

\subsubsection{Rank Fusion Strategy}

Link prediction is a list ranking problem. Rank fusion can be exploited
to boost the performance. A few simple rank fusion methods can be
applied to score-based KGE methods.  For example, we can take the
maximum, minimum, median, sum, and L2 distance of candidates' ranks.
They are denoted by CombMAX, CombMIN, CombMEDIAN, CombSUM, and Euclidean
in Table \ref{tab:rank_fusion}, respectively.  Three advanced rank
fusion methods are also considered and included in the table.  They are:
Borda Count \cite{van2000variants}, Reciprocal Rank Fusion (RRF)
\cite{cormack2009reciprocal}, and RBC (Rank Biased Centroid)
\cite{bailey2017retrieval}.  Borda Count awards points to candidates
based on their positions in an individual preference list, where the top
candidate gets the most points and the last candidate gets the least
points.  RRF aggregates the reciprocal rank to discount the importance
of lower-ranked candidates. The factor $k$ in the table mitigates the
impact of high rankings by outliers.  RBC discounts the weights of
lower-ranked candidates using a geometric distribution.  The
mathematical formulas of all rank fusion functions are given in the
second column of Table \ref{tab:rank_fusion}, where $R_i$ is the rank of
the $i$-th base model (or variant), $1\leq i \leq n$, $e\in E$
represents an entity in the entity set, and $k$ and $\phi$ are
hyper-parameters. 

\begin{table}
\caption{A list of rank fusion functions under consideration.}\label{tab:rank_fusion}
\begin{adjustbox}{width=0.8\columnwidth,center}
\begin{tabular}{cc}  \hline
Name & Function  \\  \hline
CombMAX & $\mathlarger{\max\{R_1(e),\cdots,R_n(e)\}}$\\
CombMIN & $\mathlarger{\min\{R_1(e),\cdots,R_n(e)\}}$\\
CombMEDIAN & $\mathlarger{\text{median}\{R_1(e),\cdots,R_n(e)\}}$\\
CombSUM & $\mathlarger{\sum}_{i=1}^n R_i(e)$\\
Euclidean & $\sqrt{R_1(e)^2+\cdots+R_n(e)^2}$\\ 
Borda Count & $\mathlarger{\sum_{i=1}^n\frac{|E|-R_i(e)+1}{|E|}}$\\
RRF \cite{cormack2009reciprocal} & $\mathlarger{\sum_{i=1}^n\frac{1}{k+R_i(e)}}$\\
RBC \cite{bailey2017retrieval} & $\mathlarger{\sum}_{i=1}^n(1-\phi)\phi^{R_i(e)-1}$\\ \hline
\end{tabular}
\end{adjustbox}
\end{table}

\subsection{Optimization} 

By following RotatE's negative sampling loss and the self-adversarial
training strategy, we choose the following loss function of CompoundE3D
\begin{eqnarray}
 L_{\textnormal{KGE}}=&& - \log\sigma(\zeta_1-f_r(h, t))  \\
&& -\sum_{i=1}^np(h'_i,r,t'_i) \log\sigma(f_r(h'_i,t'_i)-\zeta_1), \nonumber
\end{eqnarray}
where $\sigma$ is the sigmoid function, $\zeta_1$ is a preset margin
hyper-parameter, $(h'_i,r,t'_i)$ is the $i$-th negative triple, and
$p(h'_i,r,t'_i)$ is the probability of drawing negative triple
$(h'_i,r,t'_i)$. Given a positive triple, $(h_i,r,t_i)$,
the negative sampling distribution is
\begin{equation}
p(h'_j,r,t'_j|\{(h_i,r,t_i)\})=\frac{\exp \alpha_1 f_r(h'_j, t'_j)}
{\sum_i \exp \alpha_1 f_r(h'_i, t'_i)},
\end{equation}
where $\alpha_1$ is the temperature in the softmax function.

\section{Experiments}\label{3Dsec:experiments}

\begin{table*}[ht]
\centering
\caption{Statistics of four link prediction datasets.}\label{tab:dataset_statistics}
\begin{adjustbox}{width=0.85\textwidth,center}
\begin{tabular}{cccccccc} \hline
\textbf{Dataset} & \textbf{\#Entities} & \textbf{\#Relations} 
& \textbf{\#Training} & \textbf{\#Validation} & \textbf{\#Test} & \textbf{Ave. Degree} \\ \hline
DB100K & 99,604 & 470 & 597,572 & 50,000 & 50,000 & 12 \\
ogbl-wikikg2 & 2,500,604 & 535 & 16,109,182 & 429,456 & 598,543 & 12.2 \\
YAGO3-10 & 123,182 & 37 & 1,079,040 & 5,000 & 5,000 & 9.6 \\
WN18RR & 40,943 & 11 & 86,835 & 3,034 & 3,134 & 2.19 \\     \hline
\end{tabular}
\end{adjustbox}
\end{table*}

\begin{table*}[ht!]
\centering
\caption{The search space of six hyper-parameters.}\label{tab:hyperparameter_space}
\begin{tabular}{c|c|c|c|c}\hline
\textbf{Dataset} &  \textbf{DB100K} & \textbf{ogbl-wikikg2} & \textbf{YAGO3-10} 
& \textbf{WN18RR} \\  \hline
\textbf{Dim} & $\{150, 300, 450, 600\}$ & $\{90, 150, 180, 240, 300\}$ & $\{450, 600, 750, 900\}$ 
& $\{180, 240, 360, 480, 600\}$ \\
\textbf{\textit{lr}} & $\{2,3,4,5,6,7,8,9\}\times 10^{-5}$ & $\{0.0005,0.001,0.005,0.01\}$ 
& $\{3,4,5,6,7\}\times 10^{-4}$ & $\{4,5,6,7,8\}\times 10^{-4}$ \\
\textbf{\textit{B}} & $\{256, 512, 1024, 2048\}$ & $\{2048,4096,8192\}$ 
& $\{512, 1024, 2048, 4096\}$ & $\{512, 1024, 2048, 4096\}$ \\
\textbf{\textit{N}} & $\{256, 512, 1024, 2048\}$ & $\{125, 250, 500\}$ 
& $\{256, 512, 1024, 2048\}$ & $\{256, 512, 1024, 2048\}$ \\
$\boldsymbol{\zeta}$ & $\{4,5,6,7,8,9,10,11,12,13\}$ & $\{5,6,7,8,9\}$ 
& $\{11,12,13,13.1,13.3,13.5\}$ & $\{5,6,7,8,9\}$ \\
$\boldsymbol{\alpha}$ & $\{0.5,0.7,0.9,1.0,1.2\}$ & $\{0.5,1.0\}$ 
& $\{0.8,0.9,1.0,1.1,1.2\}$ & $\{0.5,0.7,0.9,1.0,1.2\}$ \\ \hline
\end{tabular}
\end{table*}

\begin{table*}[ht]
\centering
\caption{Optimal configurations for link prediction tasks, where
$\textit{B}$ and $\textit{N}$ denote the batch size and the negative
sample size, respectively.}\label{tab:optimal_configuration}
\begin{adjustbox}{width=0.8\textwidth,center}
\begin{tabular}{cccccccc}\hline
\textbf{Dataset} & \textbf{CompoundE3D Variant} & \textbf{\#Dim} & \textbf{\textit{lr}} 
& \textbf{\textit{B}} & \textbf{\textit{N}} & $\boldsymbol{\zeta}$ & $\boldsymbol{\alpha}$ \\ \hline
DB100K & $\|\mathbf{S\cdot h-\hat{T}\cdot\hat{R}\cdot\hat{S}\cdot t}\|$  
& 600 & 0.00005 & 1024 & 512 & 9 & 1\\
ogbl-wikikg2 & $\|\mathbf{T\cdot h-\hat{H}\cdot t }\|$ 
& 300 & 0.001 & 8192 & 125 & 8 & 1\\
YAGO3-10 & $\left\|\mathbf{T\cdot S\cdot R\cdot h-t}\right\|$ 
& 600 & 0.0005 & 1024 & 1024 & 13.3 & 1.1\\
WN18RR & $\|\mathbf{R\cdot S\cdot T\cdot h -  t }\|$ 
& 480 & 0.00005 & 512 & 256 & 6 & 1\\ \hline
\end{tabular}
\end{adjustbox}
\end{table*}

\subsection{Experimental Setup}

\subsubsection{Datasets} 

We evaluate the link prediction performance of CompoundE3D and compare
it with several benchmarking methods on the following four KG datasets. 
\begin{itemize}
\item {\bf DB100K \cite{ding2018improving}.} It is a subset of the DBpedia KG. The dataset
contains information related to music content such as genre, band, and
musical artisits.  It is a relatively dense KG since each entity appears
in at least 20 different relations. 
\item {\bf YAGO3-10 \cite{mahdisoltani2014yago3}.} It is a subset of YAGO3, which describes
citizenship, gender, and profession of people.  YGGO3-10 contains
entities associated with at least 10 different relations. 
\item {\bf WN18RR \cite{bordes2013translating, dettmers2018convolutional}.} It is a subset of the WordNet lexical database. The
inverse relation is removed from WN18RR to avoid test leakage. 
\item {\bf Ogbl-Wikikg2 \cite{hu2020ogb}.} It is extracted from Wikipedia. It contains 2.5M entities and is the
largest one among the four selected datasets. 
\end{itemize}
The statistics of the four KG datasets are given in Table
\ref{tab:dataset_statistics}. 

\subsubsection{Evaluation Protocol} 

The commonly used evaluation protocol for the link prediction task is
explained below. For every triple $(h, r ,t)$ in the test set, we
corrupt either the head entity $h$ or tail entity $t$ to generate test
examples $(?, r ,t)$ and $(h, r ,?)$. Then, for every head candidate
that forms triple $(\hat{h}, r ,t)$ and tail candidate that forms triple
$(h, r ,\hat{t})$, we compute distance-based scoring functions
$f_r(\hat{h},t)$ and $f_r(h,\hat{t})$, respectively.  The lower score
value indicates that the generated triple is more likely to be true.
Then, we sort scores of all candidate triples in ascending order and
locate the rank of the ground truth triple. Furthermore, we evaluate the
link prediction performance under the filtered rank setting \cite{bordes2013translating} that gives
salience to unseen triple predictions since embedding models tend to
give observed triples better ranks. We adopt the Hits@$k$ and the mean
reciprocal rank (MRR) as evaluation metrics to compare the quality of
KGE models. 

\subsubsection{Hyper-parameter Search}\label{subsubsec:search}

We perform an extensive search on six hyper-parameters of CompoundE3D
with respect to different KG datasets. They are: 1) the dimension of the
embedding space ({\bf Dim}), 2) the learning rate
(\textbf{\textit{lr}}), 3) the batch size ($\textit{B}$), 4) the
negative sample size ($\textit{N}$), 5) the margin hyper-parameter
($\boldsymbol{\zeta}$), and 6) the sampling temperature
($\boldsymbol{\alpha}$). Their search values are listed in Table
\ref{tab:hyperparameter_space}.  

In the search process, we first compute scoring functions with a certain
hyper-parameter setting that allows a few variants to have decent
performance, where the number of training iterations for each variant is
set to $l=30000$.  After locating the optimal variant, we finetune
hyper-parameters under the optimal variant. The optimal configurations
are shown in Table \ref{tab:optimal_configuration}.  The Adam optimizer
\cite{kingma2014adam} is employed for all parameter tuning. For ensemble
experiments, we adopt the same optimal configuration for each base
variant model. 

\subsubsection{Other Implementation Details}\label{subsubsec:hardware}

We run experiments and perform hyper-parameter tuning on a variety of
GPUs, including Nvidia P100 (16G), V100 (32G), A100 (40G) and A40 (48G),
depending on the GPU memory requirement of a job. Typically, we request
8 CPU cores with less than 70G RAM for each job. Results of each optimal
configuration in Table \ref{tab:optimal_configuration} can be reproduced
on one single V100 for all datasets. For the WN18RR dataset, we adopt the
rotation implementation from Rotate3D \cite{gao2020rotate3d}. 

\begin{table}[ht!]
\centering
\caption{Comparison of the link prediction performance under the filtered 
rank setting for DB100k.}\label{tab:db100k_lp}%
\begin{adjustbox}{width=\columnwidth,center}
\begin{tabular}{c|cccccc}  \hline
\textbf{Datasets} & \multicolumn{4}{c}{\textbf{DB100K}} \\ \hline
Model & \textbf{MRR} & \textbf{H@1} & \textbf{H@3} & \textbf{H@10} \\ \hline
TransE \cite{bordes2013translating} & 0.111 & 0.016   & 0.164  & 0.27 \\
DistMult \cite{yang2014embedding} & 0.233 & 0.115  & 0.301  & 0.448 \\
HolE \cite{nickel2016holographic} & 0.26  & 0.182  & 0.309  & 0.411 \\
ComplEx \cite{trouillon2016complex} & 0.242 & 0.126  & 0.312  & 0.44 \\
Analogy \cite{liu2017analogical} & 0.252 & 0.142  & 0.323  & 0.427 \\
RUGE  \cite{guo2018knowledge} & 0.246 & 0.129  & 0.325  & 0.433 \\
ComplEx-NNE+AER \cite{ding2018improving} & 0.306 & 0.244  & 0.334  & 0.418 \\
SEEK \cite{xu2020seek} & 0.338 & 0.268  & 0.37    & 0.467 \\
AcrE (Parallel) \cite{ren2020knowledge}   & 0.413 & 0.314  & 0.472 & 0.588 \\
PairRE \cite{chao2021pairre} & 0.412 & 0.309 & 0.472 & 0.600 \\
TransSHER \cite{li-etal-2022-transher} & 0.431 & 0.345 & 0.476 & 0.589 \\
CompoundE \cite{ge2022compounde} & 0.405 & 0.306 & 0.461 & 0.588 \\ \hline
CompoundE3D & 0.450  & 0.373 & 0.488 & 0.594 \\  \hline
CompoundE3D RRF & \underline{0.457} & \underline{0.376} & \underline{0.497} & \underline{0.607} \\
CompoundE3D WDS & \textbf{0.462} & \textbf{0.378} & \textbf{0.506} & \textbf{0.616} \\ \hline
\end{tabular}%
\end{adjustbox}
\end{table}%

\begin{table}[t]
\begin{center}
\caption{Comparison of the link prediction performance under the filtered 
rank setting for ogbl-wikikg2.}\label{tab:wikikg2_lp}
\begin{adjustbox}{width=0.8\columnwidth,center}
\begin{tabular}{c|c|c|c} \hline
\textbf{Datasets} & \multicolumn{3}{c}{\textbf{ogbl-wikikg2}} \\ \hline
\multirow{2}{*}{\textbf{Metrics}} & \multirow{2}{*}{\textbf{Dim}} 
& \textbf{Valid} & \textbf{Test}\\
& & \textbf{MRR} & \textbf{MRR} \\  \hline
AutoSF+NodePiece & 100 & 0.5806 & 0.5703 \\
ComplEx-N3-RP & 100 & 0.6701 & 0.6481  \\
TransE \cite{bordes2013translating} & 500 & 0.4272 & 0.4256 \\
DistMult \cite{yang2014embedding} & 500 & 0.3506 & 0.3729 \\
ComplEx \cite{trouillon2016complex} & 250 & 0.3759 & 0.4027 \\
RotatE \cite{sun2018rotate} & 250 & 0.4353 & 0.4353 \\
Rotate3D \cite{gao2020rotate3d} & 100 & 0.5685 & 0.5568 \\
PairRE \cite{chao2021pairre} & 200 & 0.5423 & 0.5208 \\
TripleRE \cite{yu2022triplere} & 200 & 0.6045 & 0.5794\\ 
CompoundE \cite{ge2022compounde} & 100 & 0.6704 & 0.6515 \\ \hline
\multirow{3}{*}{CompoundE3D} & 90 & 0.6994 & 0.6826 \\
& 180 & \underline{0.7146} & \underline{0.6962} \\
& 300 & \textbf{0.7175} & \textbf{0.7006} \\  \hline
\end{tabular}
\end{adjustbox}
\end{center}
\end{table}
 
\begin{table}[ht!]
\begin{center}
\caption{Comparison of the link prediction performance under the filtered 
rank setting for YAGO3-10.}\label{tab:yago3_lp}
\begin{adjustbox}{width=\columnwidth,center} 
\begin{tabular}{c|cccc}   \hline
\textbf{Datasets} & \multicolumn{4}{c}{\textbf{YAGO3-10}} \\ \hline
\textbf{Metrics} & \textbf{MRR} & \textbf{Hit@1} & \textbf{Hit@3} & \textbf{Hit@10} \\ \hline
DistMult \cite{yang2014embedding} & 0.34 & 0.24 & 0.38 & 0.54 \\
ComplEx \cite{trouillon2016complex} & 0.36 & 0.26 & 0.4 & 0.55 \\
DihEdral \cite{xu2019relation} & 0.472 & 0.381 & 0.523 & 0.643 \\
ConvE \cite{dettmers2018convolutional} & 0.44 & 0.35 & 0.49 & 0.62 \\
RotatE \cite{sun2018rotate} & 0.495 & 0.402 & 0.55 & 0.67 \\
InteractE \cite{vashishth2020interacte} & 0.541 & 0.462 & - & 0.687 \\
HAKE \cite{zhang2020learning} & \underline{0.545} & 0.462 & 0.596 & 0.694 \\
DensE \cite{lu2022dense} & 0.541 & \textbf{0.465} & 0.585 & 0.678 \\
Rot-Pro \cite{song2021rot} & 0.542	& 0.443	& 0.596	& 0.699 \\	
CompoundE \cite{ge2022compounde} & 0.477 & 0.376 & 0.538 & 0.664 \\ \hline
CompoundE3D & 0.542 & 0.450 & 0.602 & 0.701 \\ \hline
CompoundE3D RRF & 0.541 & 0.446 & \underline{0.607} & \textbf{0.707} \\
CompoundE3D WDS & \textbf{0.551} & \underline{0.463} & \textbf{0.608} & \underline{0.703} \\ \hline
\end{tabular}
\end{adjustbox}
\end{center}
\end{table}

\begin{table*}[t]
\centering
\caption{Comparison of different weighted-distances-sum (WDS) strategies for DB100K 
and YAGO3-10.}\label{tab:db100k_yago_weight}
\begin{adjustbox}{width=0.9\textwidth,center}
\begin{tabular}{c|cccc|cccc}\hline
\textbf{Datasets} & \multicolumn{4}{c|}{\textbf{DB100K}} & \multicolumn{4}{c}{\textbf{YAGO3-10}}
\\ \hline
\textbf{Strategies} & \textbf{MRR} & \textbf{H@1} & \textbf{H@3} & \textbf{H@10} & \textbf{MRR} 
& \textbf{H@1} & \textbf{H@3} & \textbf{H@10} \\ \hline
Learnable Weights & \textbf{0.462} & \textbf{0.378} & \textbf{0.506} & \textbf{0.616} 
& 0.545 & 0.451 & 0.586 & 0.696\\
Uniform Weights   & 0.460 & 0.376 & 0.503 & 0.614  & \textbf{0.551} & \textbf{0.463} 
& \textbf{0.608} & \textbf{0.703}\\
Geometric Weights & 0.446 & 0.348 & 0.503 & 0.618 & 0.531 & 0.439 & 0.580 & 0.691 \\ \hline
\end{tabular}%
\end{adjustbox}
\end{table*}%

\begin{table*}[ht!]
  \begin{center}
  \caption{Performance comparison of different rank fusion methods for DB100K and YAGO3-10.}
  \label{tab:rank_fusion_db100k_yago}
  \begin{adjustbox}{width=\textwidth,center}
    \begin{tabular}{c|cccc|cccc} 
      \hline
      \textbf{Datasets} & \multicolumn{4}{c|}{\textbf{DB100K}} & \multicolumn{4}{c}{\textbf{YAGO3-10}} \\
      \hline
      \textbf{Aggregation Function} & \textbf{MRR} & \textbf{Hit@1} & \textbf{Hit@3} & \textbf{Hit@10} & \textbf{MRR} & \textbf{Hit@1} & \textbf{Hit@3} & \textbf{Hit@10}\\
      \hline
      CombMAX & 0.455 & 0.375 & 0.496 & 0.603 & 0.536 & 0.440 & 0.600 & 0.701 \\
      CombMIN & 0.452 & 0.369 & 0.4955 & 0.606 & 0.527 & 0.427 & 0.597 & 0.702 \\
      CombMEDIAN & 0.456 & 0.376 & 0.497 & 0.606 & 0.541 & 0.445 & 0.605 & 0.705 \\
      CombSUM & 0.456 & 0.376 & 0.4969 & 0.6060 & 0.540 & 0.446 & 0.606 & 0.704 \\
      Euclidean & 0.455 & 0.375 & 0.496 & 0.605 & 0.540 & 0.445 & 0.603 & 0.702 \\
      Borda & 0.456 & 0.376 & 0.497 & 0.606 & 0.540 & 0.446 & 0.606 & 0.704 \\
      RRF & \textbf{0.457} & \textbf{0.376} & \textbf{0.497} & \textbf{0.607} & \textbf{0.541} & \textbf{0.446} & \textbf{0.607} & \textbf{0.707} \\
      RBC & 0.456 & 0.376 & 0.497 & 0.606 & 0.540 & 0.445 & 0.604 & 0.703 \\
      \hline
    \end{tabular}
    \end{adjustbox}
  \end{center}
\end{table*}

\begin{table}[ht!]
\begin{center}
\caption{Ablation study on CompoundE3D for DB100K.}\label{tab:db100k_ablation}
\begin{adjustbox}{width=\columnwidth,center}
\begin{tabular}{c|cccc} \hline
\textbf{Datasets} & \multicolumn{4}{c}{\textbf{DB100K}} \\ \hline
\textbf{Variant} & \textbf{MRR} & \textbf{Hit@1} & \textbf{Hit@3} & \textbf{Hit@10}\\ \hline
$\|\mathbf{S\cdot h-\hat{S}\cdot t }\|$ & 0.417 & 0.323 & 0.471 & 0.590 \\
$\|\mathbf{S\cdot h-\hat{R}\cdot\hat{S}\cdot t}\|$ & 0.447 & 0.364 & 0.492 & 0.600\\
$\|\mathbf{S\cdot h-\hat{T}\cdot\hat{R}\cdot\hat{S}\cdot t}\|$ & 0.450  & 0.373 & 0.488 & 0.594\\ \hline
\end{tabular}
\end{adjustbox}
\end{center}
\end{table}

\begin{table}[ht!]
\begin{center}
\caption{Ablation study on CompoundE3D for YAGO3-10.}\label{tab:yago3_ablation}%
\begin{adjustbox}{width=\columnwidth,center}
\begin{tabular}{c|cccc} \hline
\textbf{Datasets} & \multicolumn{4}{c}{\textbf{YAGO3-10}} \\ \hline
\textbf{Metrics} & \textbf{MRR} & \textbf{Hit@1} & \textbf{Hit@3} & \textbf{Hit@10} \\ \hline
$\left\|\mathbf{R\cdot h-t}\right\|$ & 0.496 & 0.402 & 0.547 & 0.676 \\
$\left\|\mathbf{S\cdot R\cdot h-t}\right\|$ & 0.501 & 0.404 & 0.554 & 0.680 \\
$\left\|\mathbf{T\cdot S\cdot R\cdot h-t}\right\|$ & 0.542 & 0.450 & 0.602 & 0.701 \\ \hline
\end{tabular}
\end{adjustbox}
\end{center}
\end{table}

\subsection{Experimental Results}\label{subsec:linkprediction}

\subsubsection{Performance Evalution}

We compare the link prediction performance of a few benchmarking KGE
methods with that of CompoundE3D using the optimal configuration given
in Table \ref{tab:optimal_configuration}.  The performance benchmarking
results for DB100K, ogbl-wikikg2, and YAGO3-10 datasets are shown,
respectively, in Table \ref{tab:db100k_lp}, Table \ref{tab:wikikg2_lp},
and Table \ref{tab:yago3_lp}. Furthermore, the best and the second-best
results in each column are indicated by the boldface font and with an
underline, respectively.  CompoundE3D has significant performance
improvement over CompoundE and other recent models. We see a clear
advantage of CompoundE3D by including more affine operators and
extending affine transformations from 2D to 3D in the new framework,

To verify the effectiveness of model ensembles, we examine two different
ensemble strategies for DB100K and YAGO3-10 datasets.
\begin{itemize}
\item For the DB100K dataset, we select the best two performing variants.
They are $\|\mathbf{S\cdot h-\hat{R}\cdot\hat{S}\cdot t}\|$ and
$\|\mathbf{S\cdot h-\hat{T}\cdot\hat{R}\cdot\hat{S}\cdot t}\|$ 
\item For the YAGO3-10 dataset, we select the best three performing variants.
They are $\left\|\mathbf{T\cdot S\cdot R\cdot h-t}\right\|$,
$\left\|\mathbf{S\cdot R\cdot T\cdot h-t}\right\|$, and
$\left\|\mathbf{S\cdot T\cdot R\cdot h-t}\right\|$. 
\end{itemize}

\subsubsection{Model Ensembles}

As discussed in Sec. \ref{subsec:fusion}, we have two strategies to
conduct model ensembles: weighted-distances-sum (WDS) and rank fusion.
Among the three WDS strategies, the learnable weight strategy is the
most effective one for DB100K while the uniform weight performs the best
for YAGO3-10. We use CompoundE3D WDS to denote the best WDS scheme in
Tables \ref{tab:db100k_lp} and \ref{tab:yago3_lp} and document the
performance of other weighting strategies in Table
\ref{tab:db100k_yago_weight}.  Among all eight rank fusion strategies,
we observe that reciprocal rank fusion (RRF) is the most effective one
for both DB100K and YAGO3-10.  Thus, we use CompoundE3D RRF to denote
the best rank fustion scheme in Tables \ref{tab:db100k_lp} and
\ref{tab:yago3_lp}, and document the performance of other rank fushion
strategies in Table \ref{tab:rank_fusion_db100k_yago}. 

\subsubsection{Effectiveness of Beam Search} 

\begin{figure}[ht!]
\centering
\includegraphics[width=\columnwidth]{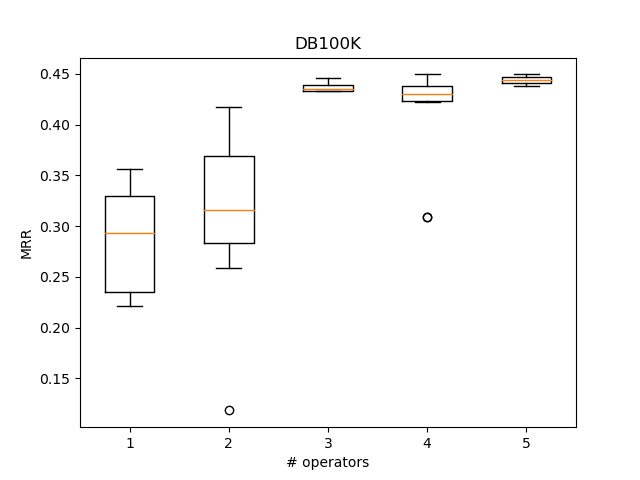}
\caption{The distribution of the MRR performance versus the operator number 
of various model variants for the DB100K dataset.} \label{fig:boxplot_db100k}
\end{figure}

\begin{figure}[ht!]
\centering
\includegraphics[width=\columnwidth]{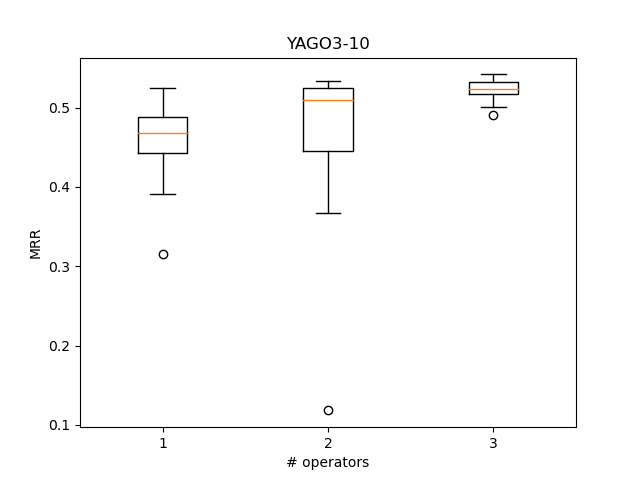}
\caption{The distribution of the MRR performance versus the operator number 
of various model variants for the YAGO3-10 dataset.} \label{fig:boxplot_yago}
\end{figure}

We conduct ablation studies on DB100K and YAGO3-10 datasets to shed
light on the effects of different transformation operators on model
performance. We begin with the variant of the simplest configuration and
add additional operators at each stage. Good simple models that lead to
optimal variants and their performance numbers are reported in Tables
\ref{tab:db100k_ablation} and \ref{tab:yago3_ablation}.  Furthermore, we
visualize the distribution of the MRR performance as more operators are 
added with respect to DB100K and YAGO-3 in Figs.  \ref{fig:boxplot_db100k} 
and \ref{fig:boxplot_yago}, respectively. To interpret box plots, yellow bar represents the median, box represents the interquantile range, two end-bars denote the lower and upper whiskers, and lastly dots are outliers. They both show the effectiveness of the proposed beam search algorithm.

\begin{figure*}[ht!]
\centering
\includegraphics[width=0.9\textwidth]
{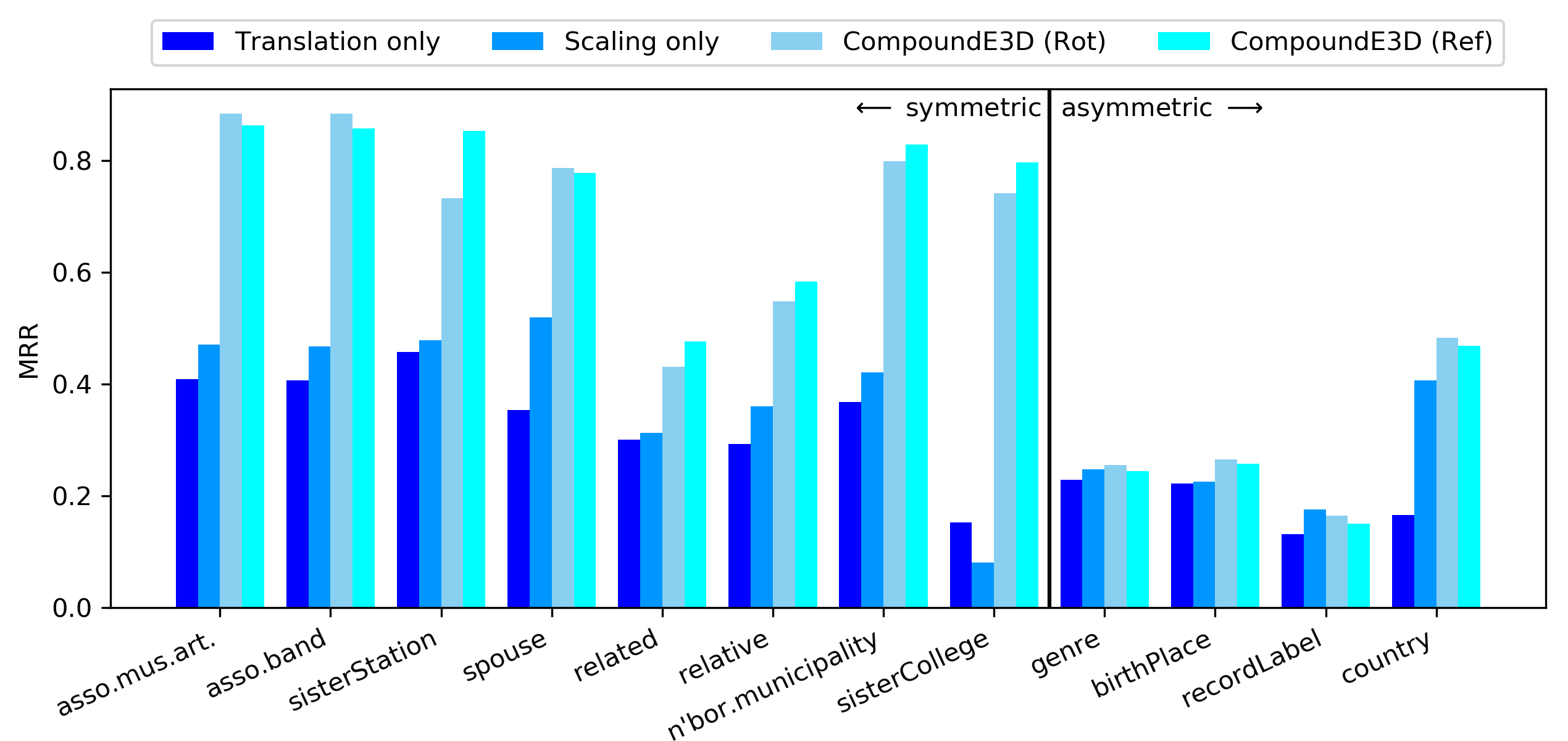}
\caption{Effects of rotation and reflection operators on symmetric relations.}
\label{fig:rotation_reflection_effect}
\end{figure*}

\begin{figure}[ht!]
\centering
\includegraphics[width=0.9\columnwidth]{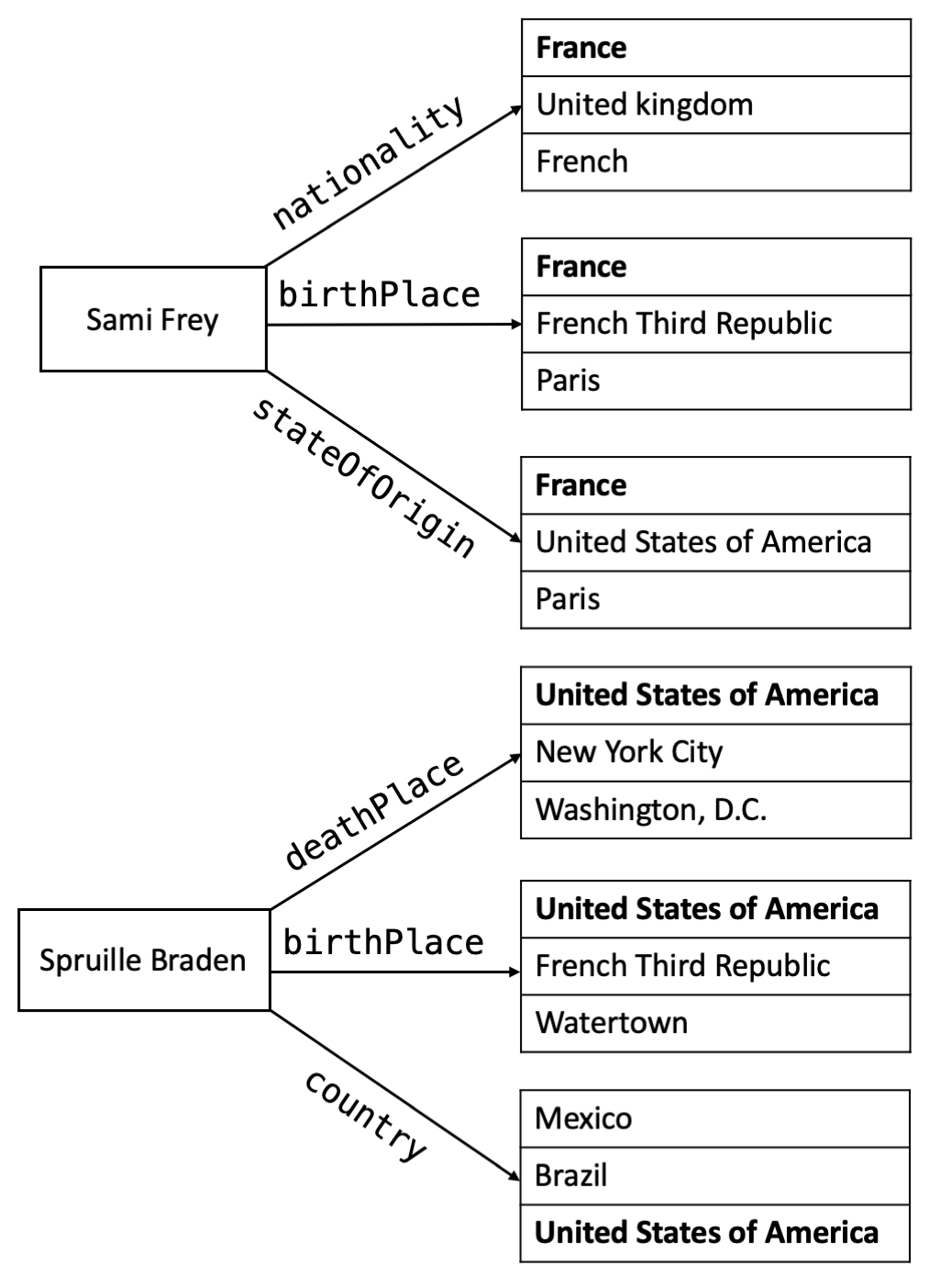}
\caption{Illustration of CompoundE3D's capability in multiplicity modeling.}
\label{fig:multiplicity_example}
\end{figure}

\begin{figure}[ht!]
\centering
\includegraphics[width=\columnwidth]{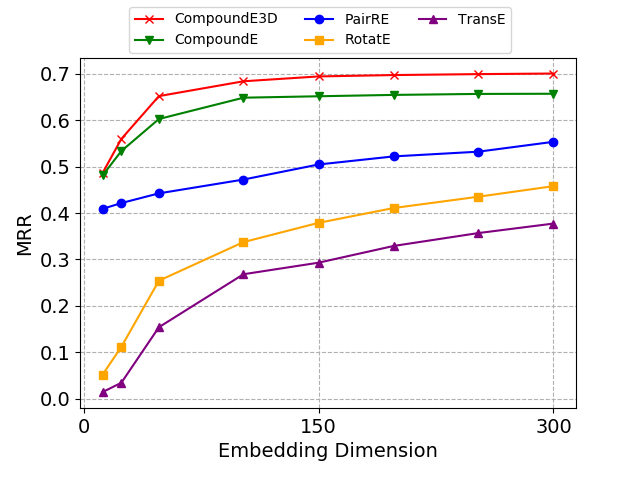}
\caption{Comparing different model's MRR performance metric across different dimensions.}
\label{fig:mrr_dim}
\end{figure}

\begin{table*}[ht!]
\begin{center}
\caption{Comparison of filtered MRR performance on each relation type of WN18RR.}
\label{tab:wn18rr_detail}
\begin{adjustbox}{width=0.99\textwidth,center}
\begin{tabular}{c|c|cc|cccc} 
\hline
\textbf{Relations} & \textbf{Types} & $\text{Khs}_{G_r}$ & $\xi_{G_r}$ & \textbf{TransE} 
& \textbf{RotatE} & \textbf{CompoundE} & \textbf{CompoundE3D}\\\hline
similar to & 1-to-1 & 0.07 & -1.00 & 0.294 & \textbf{1.000} & \textbf{1.000} & \textbf{1.000}\\
verb group & 1-to-1 & 0.07 & -0.50 & 0.363 & 0.961 & \textbf{0.974} & 0.898\\ \hline
member meronym & 1-to-N & 1.00 & -0.50 & 0.179 & \textbf{0.259} & 0.230 & 0.246\\
has part & 1-to-N & 1.00 & -1.43 & 0.117 & 0.200 & 0.190 & \textbf{0.202}\\
member of domain usage & 1-to-N & 1.00 & -0.74 & 0.113 & 0.297 & 0.332 & \textbf{0.378}\\
member of domain region & 1-to-N & 1.00 & -0.78 & 0.114 & 0.217 & 0.280 & \textbf{0.413}\\ \hline
hypernym & N-to-1 & 1.00 & -2.64 & 0.059 & 0.156 & 0.155 & \textbf{0.182}\\
instance hypernym & N-to-1 & 1.00 & -0.82 & 0.289 & 0.322 & 0.337 & \textbf{0.356}\\
synset domain topic of & N-to-1 & 0.99 & -0.69 & 0.149 & 0.339 & 0.367 & \textbf{0.396} \\ \hline
also see & N-to-N & 0.36 & -2.09 & 0.227 & 0.625 & \textbf{0.629} & 0.622\\
derivationally related form & N-to-N & 0.07 & -3.84 & 0.440 & 0.957 & 0.956 & \textbf{0.959}\\ \hline
\end{tabular}
\end{adjustbox}
\end{center}
\end{table*}

\begin{table}[ht]
\centering
\caption{Complexity comparison of KGE models on ogbl-wikikg2 under a similar testing MRR.}\label{tab:complexity}
    \begin{adjustbox}{width=0.6\columnwidth,center}
    \begin{tabular}{lc} \hline
        \textbf{Models}  & \textbf{No. of Parameters} \\ \hline
        TransE \cite{bordes2013translating} &  1,251M \\
        DistMult \cite{yang2014embedding} &  1,251M \\
        ComplEx \cite{trouillon2016complex} &  1,251M \\
        ComplEx-RP \cite{chenrelation} & 250.1M \\
        RotatE \cite{sun2018rotate} & 500M \\ 
        RotatE3D \cite{gao2020rotate3d} &  750.4M \\ 
        PairRE \cite{chao2021pairre} &  500M \\
        CompoundE \cite{ge2022compounde} &  250.1M \\ 
        CompoundE3D &  \textbf{225.2M} \\ \hline
    \end{tabular}
    \end{adjustbox}
\end{table}

\subsubsection{Modeling of Symmetric Relations} 

Rotation and reflection are isometric operations. As stated in
\cite{sun2018rotate, zhang2022knowledge}, their 2D versions can handle
symmetric relations well in some cases. It is our conjecture that the
same property holds for their corresponding 3D operators. To check it,
we perform ablation studies and evaluate the base scoring functions of
those with only translation and scaling versus those with rotation and
reflection as well.  The MRR performance numbers of different model
variants for symmetric and asymmetric relations in DB100K are compared
in Fig.  \ref{fig:rotation_reflection_effect}. In this figure, we choose
the most frequently observed relation types for meaningful comparison.
As expected, rotation and reflection operators indeed bring more
significant performance improvement on symmetric relations than
asymmetric relations. This supports our conjecture that rotation and
reflection operators are intrinsically advantageous for the modeling of
symmetric relations. 

\subsubsection{Modeling of Multiplicity} 

Multiplicity is the scenario where multiple relations co-exist between
two entities; namely, triples $(h,r_1,t), \dots, (h,r_n,t)$ hold
simultaneously. Generally, it is challenging to model multiplicity in
traditional KGE models due to their limited power in relational
modeling.  In contrast, CompoundE3D is capable of modeling multiplicity
relation patterns well since it can use multiple distinct sets of
transformations that map from the head to the tail.  We present two
examples to illustrate CompoundE3D's capability in modeling multiplicity
relations in Fig.  \ref{fig:multiplicity_example}. They are taken from
the actual link prediction examples in DB100K. There are three different
relations held for a fixed (head, tail) pair. The top three tail
predictions for each relation in the two examples are shown in the
figure.  We see that CompoundE3D can handle multiplicity well due to its
rich set of variants. 

\subsubsection{Modeling of Hierarchical Relations} 

We would like to investigate CompoundE3D's capability in modeling
hierarchical relations.  WN18RR offers a representative dataset
containing hierarchical relations. Two metrics can be used to measure
the hierarchical behavior of relations \cite{chami2020low}: 1) the
Krackhardt score denoted by $\text{Khs}_{G_r}$, and 2) the curvature
estimate denoted by $\xi_{G_r}$. If relation $r$ has a high
$\text{Khs}_{G_r}$ score and a low $\xi_{G_r}$ score, then it has a
stronger hierarchical behavior, and vice versa.  We compare the filtered
MRR performance of different baseline models, such as TransE, RotatE,
CompoundE (2D version), and CompoundE3D in Table
\ref{tab:wn18rr_detail}.  In the same table, we also list the
$\text{Khs}_{G_r}$ and $\xi_{G_r}$ values for each relation to see
whether it has a stronger hierarchical behavior.  We see from the table
that CompoundE and CompoundE3D have better performance than TransE and
RotatE in almost all relations. Furthermore, CompoundE3D outperforms
CompoundE in all hierarchical relations except ``member meronym". This
result indicates that CompoundE3D can model hierarchical relations more
effectively than CompoundE by including more diverse 3D transformations. 

\subsubsection{Model Efficiency}

It is important to investigate the relationship between the model
performance and the model dimension. The model dimension reflects memory
and computational complexities.  To illustrate the advantage of
CompoundE3D over prior models across a wide range of embedding
dimensions, we plot the MRR performance of link prediction on the
Wikikg2 dataset in Fig. \ref{fig:mrr_dim}, where the dimension values
are set to 12, 24, 48, 102, 150, 198, 252, and 300.  We see from the
figure that CompoundE3D consistently outperforms all benchmarking models
in all dimensions.  Furthermore, we analyze the complexity of different
KGE models in terms of the number of free parameters.  Table
\ref{tab:complexity} compares the number of free parameters of different
KGE models for the ogbl-wikikg2 dataset.

We refer to the ogbl-wikikg2 leaderboard when reporting the number of free parameters used by baseline models. The reported number of parameters for CompoundE3D is when the embedding dimension is set to 90. As shown in Fig. \ref{fig:mrr_dim} and Table
\ref{tab:complexity}, CompoundE3D offers the best performance among all
benchmarking models while having the smallest number of free parameters. 

\section{Conclusion and Future Work}\label{3Dsec:conclusion}

A novel and effective KGE model based on composite affine
transformations in the 3D space, named CompoundE3D, was proposed in this
work. A beam search procedure was devised to build a desired KGE from
the simplest configuration to more complicated ones.  The ensemble of
the top-$k$ model variants was also explored to further boost link
prediction performance. Extensive experimental results were provided to
demonstrate the superior performance of CompoundE3D. We conducted
ablation studies to assess the effect of each operator and performed
case studies to shed light on the modeling power of CompoundE3D for
several relation types such as multiplicity, symmetric relations, and
hierarchical relations. 

As to future research directions, it will be interesting to explore the
effectiveness of CompoundE3D in other important KG problems such as
entity typing \cite{ge2022core} and entity alignment
\cite{ge2023typeea}. Besides, research on performance boosting in
low-dimensional embedding space is valuable in practical real-world
applications and worth further investigation. 

\section*{Acknowledgment}

The authors acknowledge the Center for Advanced Research Computing (CARC) at the
University of Southern California for providing computing resources that
have contributed to the research results reported within this publication. URL: \url{https://carc.usc.edu}.

\bibliography{abrv,conf_abrv,reference}
\bibliographystyle{IEEEtran}

\begin{IEEEbiography}[{\includegraphics[width=1in,height=1.25in,clip,keepaspectratio]
{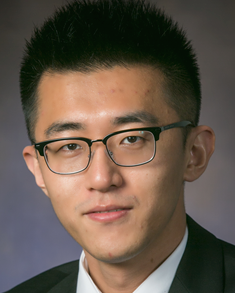}}]
{Xiou Ge (S'14)} received the B.S. and M.S. degree in electrical and
computer engineering from University of Illinois, Urbana–Champaign,
Urbana, Illinois, in 2016 and 2018 respectively. He is working towards
Ph.D. degree in electrical and computer engineering with the Ming-Hsieh
Department of Electrical and Computer Engineering, University of
Southern California. His research interest includes knowledge graph
embedding, node classification, graph alignment, natural language
processing. 
\end{IEEEbiography}

\begin{IEEEbiography}[{\includegraphics[width=1in,height=1.25in,clip,keepaspectratio]
{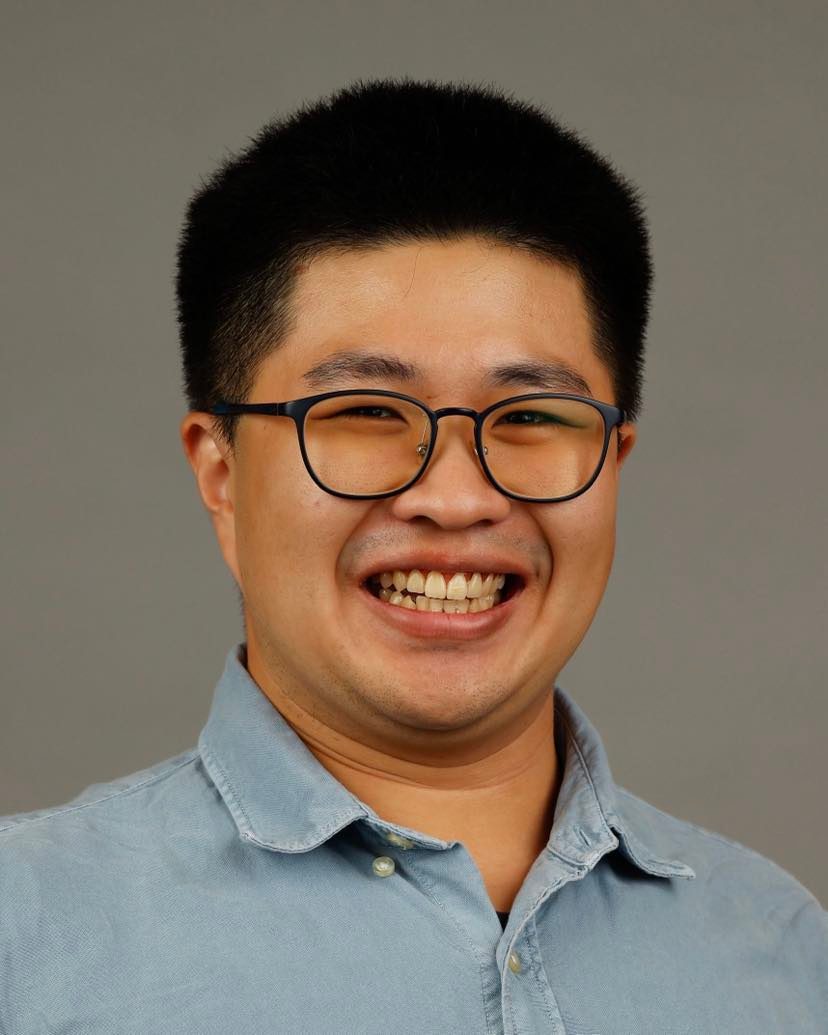}}]
{Yun-Cheng Wang (S'19)} received the B.S. degree in electrical engineering
from National Taiwan University, Taipei, Taiwan, in 2018. He received
his M.S. degree in 2020 and is working towards Ph.D. degree in
electrical and computer engineering with the Ming-Hsieh Department of
Electrical and Computer Engineering, University of Southern California.
His research interest includes knowledge graph embedding, natural
language processing. 
\end{IEEEbiography}

\begin{IEEEbiography}[{\includegraphics[width=1in,height=1.25in,clip,keepaspectratio]
{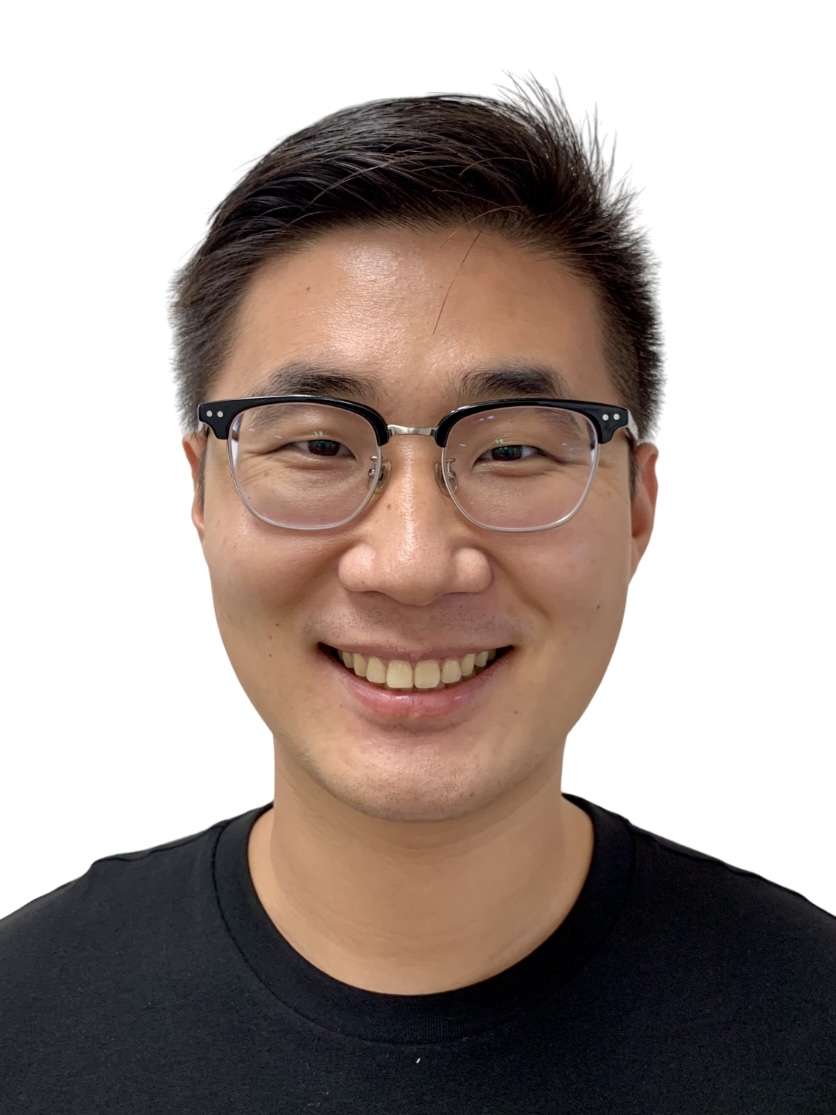}}]
{Bin Wang (S'19-M'22)} received his B.Eng. degree in electronic
information engineering from the University of Electronic Science and
Technology of China (UESTC), Chengdu, China, in June 2017, and his Ph.D.
degree in electrical engineering from the University of Southern
California (USC), Los Angeles, CA, USA, in May 2021. He is currently a
Research Fellow with the National University of Singapore (NUS),
Singapore. His research interests include representation learning, graph
learning, and dialogue summarization. 
\end{IEEEbiography}

\begin{IEEEbiography}[{\includegraphics[width=1in,height=1.25in,clip,keepaspectratio]
{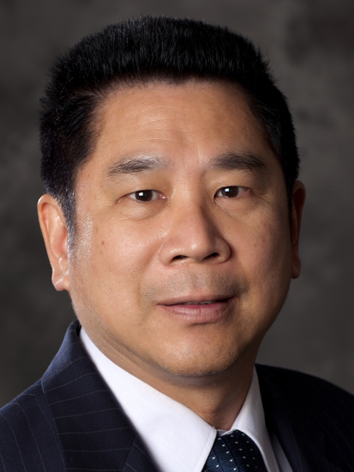}}]
{C.-C. Jay Kuo (F’99)} received the B.S. degree in electrical
engineering from National Taiwan University, Taipei, Taiwan, in 1980,
and the M.S. and Ph.D. degrees in electrical engineering from the
Massachusetts Institute of Technology, Cambridge, in 1985 and 1987,
respectively. He is the holder of the William M. Hogue Professorship in
Electrical and Computer Engineering, a Distinguished Professor of
Electrical and Computer Engineering and Computer Science, and the
Director of the USC Multimedia Communication Laboratory (MCL) at the
University of Southern California. He is the co-author of about 340
journal papers, 1000 conference papers, and 15 books. His research
interests include multimedia and green computing. He is a fellow of the
National Academy of Inventors, the American Association for the
Advancement of Science, and the International Society for Optical
Engineers, and the Association for Computing Machinery. 
\end{IEEEbiography}

\end{document}